\documentclass[lettersize,journal]{IEEEtran}
\usepackage{amsmath,amsfonts}
\usepackage{algorithmic}
\usepackage{algorithm}
\usepackage{array}
\usepackage{textcomp}
\usepackage{stfloats}
\usepackage{url}
\usepackage{verbatim}
\usepackage{graphicx}
\usepackage{cite}
\newtheorem{definition}{Definition}
\newtheorem{assumption}{Assumption}
\usepackage{amsmath}
\usepackage{amsfonts} 
\usepackage{tabularx} 

\newcommand{\indep}{\perp \!\!\! \perp} 
\usepackage{subfigure}
\usepackage{tikz}
\usetikzlibrary{automata,arrows,calc,positioning}
\usetikzlibrary{shadows,arrows,intersections,matrix,backgrounds,shapes,shapes.geometric, positioning, fit}
\usetikzlibrary{patterns,decorations.pathmorphing}
\pgfarrowsdeclare{arcsq}{arcsq}
{
	\arrowsize=0.2pt
	\advance\arrowsize by .5\pgflinewidth
	\pgfarrowsleftextend{-4\arrowsize-.5\pgflinewidth}
	\pgfarrowsrightextend{.5\pgflinewidth}
}
{
	\arrowsize=0.8pt
	\advance\arrowsize by .5\pgflinewidth
	\pgfsetdash{}{0pt} 
	\pgfsetroundjoin   
	\pgfsetroundcap    
	\pgfpathmoveto{\pgfpoint{0\arrowsize}{0\arrowsize}}
	\pgfpatharc{-90}{-140}{4\arrowsize}
	\pgfusepathqstroke
	\pgfpathmoveto{\pgfpointorigin}
	\pgfpatharc{90}{140}{4\arrowsize}
	\pgfusepathqstroke
}
\usepackage{booktabs} 

\begin{document}

\title{A Survey on Causal Reinforcement Learning}


\author{Yan Zeng, Ruichu Cai, \IEEEmembership{Senior Member, IEEE}, Fuchun Sun, \IEEEmembership{Fellow, IEEE},  Libo Huang, Zhifeng Hao, \IEEEmembership{Senior Member, IEEE}
\thanks{Yan Zeng is with the Department of Computer Science and Technology, Tsinghua University, Beijing 100084, China (e-mail: yanazeng013@tsinghua.edu.cn).}
\thanks{Ruichu Cai is with the School of Computer Science, Guangdong University of Technology, Guangzhou 510006, China, and also with the Peng Cheng Laboratory, Shenzhen 518066, China (e-mail: cairuichu@gmail.com).}
\thanks{Fuchun Sun is with the Department of Computer Science and Technology, Beijing National Research Centre for Information Science and Technology, Tsinghua University, Beijing 100084, China (Corresponding author, e-mail: fcsun@tsinghua.edu.cn).}
\thanks{Libo Huang is with Institute of Computing Technology, Chinese Academy of Sciences, Beijing, 100190, China (e-mail: www.huanglibo@gmail.com).}
\thanks{Zhifeng Hao is with the College of Science, Shantou University, Shantou 515063, China (e-mail: haozhifeng@stu.edu.cn).}
}

\markboth{Journal of \LaTeX\ Class Files,~Vol.~14, No.~8, August~2021}%
{Shell \MakeLowercase{\textit{et al.}}: A Sample Article Using IEEEtran.cls for IEEE Journals}


\maketitle

\begin{abstract}
While Reinforcement Learning (RL) achieves tremendous success in sequential decision-making problems of many domains, it still faces key challenges of data inefficiency and the lack of interpretability. Interestingly, many researchers have leveraged insights from the causality literature recently, bringing forth flourishing works to unify the merits of causality and address well the challenges from RL.
As such, it is of great necessity and significance to collate these Causal Reinforcement Learning (CRL) works, offer a review of CRL methods, and investigate the potential functionality from causality toward RL.
In particular, we divide existing CRL approaches into two categories according to whether their causality-based information is given in advance or not. We further analyze each category in terms of the formalization of different models, ranging from the Markov Decision Process (MDP), Partially Observed Markov Decision Process (POMDP), Multi-Arm Bandits (MAB), and Dynamic Treatment Regime (DTR). Moreover, we summarize the evaluation matrices and open sources, while we discuss emerging applications, along with promising prospects for the future development of CRL.
\end{abstract}


\section{Introduction}
 \IEEEPARstart{R}{einforcement} 
Learning (RL) is a general framework for an agent to learn a policy (the mapping function from states to actions) which maximizes the expected rewards in an environment~\cite{sutton2018reinforcement,henderson2018deep,mnih2015human}. 
It attempts to tackle sequential decision-making problems through the scheme of trial and error while the agent interacts with the environment. 
Due to its remarkable success in performance, it has been rapidly developed and deployed in various real-world applications, including games~\cite{mnih2013playing,jaderberg2019human,vinyals2019grandmaster}, robotics controlling~\cite{kober2013reinforcement,finn2016guided}, and recommender systems~\cite{lin2021survey,sun2021cost}, etc., gaining increasing attention from researchers of different disciplines.  

However, there are some key challenges of reinforcement learning that still need to be addressed. For instance, (i) {\it data inefficiency}. Previous methods are mostly hungry for interaction data, whereas in real-world scenarios, e.g., in medicine or healthcare~\cite{yu2021reinforcement}, only a few recorded data are available, majorly due to the costly, unethical or difficult collection procedures. (ii) {\it Lack of interpretability}. Existing methods often formalize the reinforcement learning problem through deep neural networks, which belong to the black boxes, taking sequential data as inputs and the policy as output. It is hard for them to uncover the internal relationships between states, actions, or rewards behind the data, and to offer intuitions about the policy characteristics. Such a challenge would impede the actual applications in industrial engineering.

Interestingly, causality may play an indispensable role in handling those aforementioned challenges from reinforcement learning~\cite{bareinboim2015bandits,peters2017elements}. 
Causality considers two fundamental questions~\cite{Pearl2000causality}: (1) What empirical evidence is required for legitimate inference of cause-effect relationships? This procedure of uncovering cause-effect relationships with evidence is briefly termed causal discovery. (2) Given the accepted causal information about a phenomenon, what inferences can we draw from such information, and how? Such a procedure of inferring causal effects or other interests is termed causal inference.
Causality could empower the agents to do interventions or reason counterfactually via the {\it ladder of causation}, relaxing the requirements of a large amount of training data; it also enables to characterize the world model, potentially giving interpretability to how the agents interact with the environment.

In the past decades, both causality and reinforcement learning have achieved tremendous theoretical and technical development independently, whereas they could have been reconcilably integrated with each other~\cite{Elias2020icml}. 
Bareinboim~\cite{Elias2020icml} developed a unified framework called causal reinforcement learning by putting them under the same conceptual and theoretical umbrella and provided a tutorial for introduction online; Lu~\cite{lu2018introduction} was motivated by the current development in healthcare and medicine, and thus integrated causality and reinforcement learning, introducing causal reinforcement learning as well as emphasizing its potential applicability.
Recently, a series of research related to causal reinforcement learning has been proposed, which needs a comprehensive survey on its development and applications. Thus, in this paper, we focus on providing our readers with good knowledge about concepts, categories, and practical issues of causal reinforcement learning. 

Though there existed some relative reviews, i.e., Grimbly et al.,~\cite{grimbly2021causal} surveyed on causal multi-agent reinforcement learning; Bannon et al.,~\cite{bannon2020causality} on causal effects' estimation and off-policy evaluation in batch reinforcement learning, here we consider cases but not limited to multi-agent or off-policy evaluation ones. Recently, Kaddour et al.,~\cite{kaddour2022causal} have uploaded a survey on causal machine learning on arXiv, including a chapter on causal reinforcement learning. They summarized the approaches according to different RL problems that causality could bring benefits to, e.g., causal bandits, model-based RL, off-policy policy evaluation, etc. Such a taxonomic way may not be complete or integrated, which misses some other RL problems, e.g., multi-agent RL~\cite{bannon2020causality}.
In this paper we only but completely construct a taxonomy framework for those causal reinforcement learning approaches. Our contributions of this survey paper are as follows:
\begin{itemize}
    \item We formally define causal reinforcement learning, and to the best of our knowledge, we distinguish existing approaches into two categories from the perspective of causality for the first time. The first category is based on the prior causal information, where generally such methods assume the causal structure regarding to the environment or task is given a prior from the experts, while the second category is based on the unknown causal information, where the relative causal information has to be learned for the policy.
    \item We offer a comprehensive review of current methods on each category with systematical descriptions (and sketch maps). Regarding the first category, CRL methods take the best of prior causal information in policy learning, to improve sample efficiency, causal explanation or generalization capability. Regarding CRL with unknown causal information, these methods usually contain two phases: causal information learning and policy learning, which are carried out iteratively or in turn.
    \item We further analyze and discuss applications of CRL, evaluation metrics, open sources as well as future directions~\footnote{Please see Appendix for evaluation metrics and open sources.}.
\end{itemize}

\section{Preliminary}
Here we give basic definitions and concepts in both causality and RL, which will be utilized throughout the paper.  

\subsection{Causality}
As for causality, we first define the notations and assumptions, and then introduce general methods from perspectives of causal discovery and causal inference. The former addresses the problem of identifying causal structures from purely observational data, which exploits statistical properties to tell the cause-effect relations between variables of interest; while the latter aims at inferring causal effects or other statistical interests when the causal relations are completely or partially known.

\begin{itemize}
    \item {\textit{Definitions and assumptions}}
\end{itemize}
%
\begin{definition}[\textbf{Structural Causal Model} (SCMs)~\cite{Pearl2000causality}]
Given a set of equations in Eq.\eqref{eq:scm}, if each equation represents an autonomous mechanism and each mechanism determines the value of only one distinct variable, then the model is called a structural causal model or causal model for short,
\begin{equation} \label{eq:scm}
    x_i = f_i(pa_i, u_i), i=1,...,n,
\end{equation}
where $x_i, u_i$ are the $i^{th}$ random variable and its error variable, respectively. $f_i$ stands for the generation mechanism while $pa_i$ is a set of $x_i$'s parental variables, i.e., those immediate causes of $x_i$. 
\end{definition}
Given a SCM with two random variables, which satisfies,
\begin{equation} \label{eq:scm2}
    x_1 = u_1, 
    x_2 = f_2(x_1, u_2),
\end{equation}
then $x_1 \to x_2$ is referred to as a causal graph~\cite{peters2017elements}. Here, $x_1$ is the cause or parental parent of $x_2$. Every causal model can be associated with a directed graph.

\begin{definition}[\textbf{Rubin Causal Model}~\cite{rubin1974estimating,sekhon2008neyman}]
Rubin causal model involves an observational dataset of $\lbrace Y_i, T_i, X_i \rbrace$, where $Y_i$ denotes a potential outcome of unit $i$; $T_i\in \lbrace 0, 1\rbrace$ is an indicator variable for taking a treatment or not; and $X_i$ is a set of covariates. 
\end{definition}

The Rubin causal model is also well-known as the potential outcome framework or the Neyman-Rubin potential outcomes. 
Since an individual unit cannot take different treatments simultaneously, but rather only one treatment at a time, it is impossible to obtain both potential outcomes and one has to estimate the missing one. With potential outcomes, the Rubin causal model aims at estimating the treatment effect.

\begin{definition}[\textbf{Treatment effect}]
Treatment effects can be evaluated at the population, treated group, subgroup, and individual levels. 
As for the population level, the Average Treatment Effect (ATE) is defined as
\begin{equation}
    \mathrm{ATE} = \mathbb{E}[Y(T=1) - Y(T=0)],
\end{equation}
where $Y(T=1)$ and $Y(T=0)$ represent the potential outcomes with and without treatment, respectively. 
As for the treated group level, the Average Treatment effect on the Treated group (ATT) is defined as
\begin{equation}
    \mathrm{ATT} = \mathbb{E}[Y(T=1)\mid T=1] - \mathbb{E}[Y(T=0)\mid T=1],
\end{equation}
where $Y(T=1)\mid T=1$ and $Y(T=0)\mid T=1$ represent the potential outcomes with and without treatment, respectively, in the treated group. 
As for the subgroup level, the Conditional Average Treatment Effect (CATE) is defined as
\begin{equation}
    \mathrm{CATE} = \mathbb{E}[Y(T=1)\mid X=x] - \mathbb{E}[Y(T=0)\mid X=x],
\end{equation}
where $Y(T=1)\mid X=x$ and $Y(T=0)\mid X=x$ represent the potential outcomes with and without treatment, respectively, of the subgroup with $X=x$. 
As for the individual level, the Individual Treatment Effect (ITE) is defined as
\begin{equation}
    \mathrm{ITE}_i = Y_i(T=1) - Y_i(T=0),
\end{equation}
where $Y_i(T=1)$ and $Y_i(T=0)$ represent the potential outcomes of the unit $i$ with and without treatment, respectively. \end{definition}

\begin{definition}[\textbf{Confounder}~\cite{glymour2019review}]
A confounder in a causal graph is the unobserved direct common cause of two observed variables. Especially, confounders in the potential outcome framework are those variables that affect both the treatment and the outcome.
\end{definition}

\begin{definition}[\textbf{Instrumental Variables (IVs)}~\cite{Pearl2000causality}]
An variable $Z$ is an IV relative to the pair $(T, Y)$, if it satisfies the following two conditions: i) $Z$ is independent of all variables (including error terms) that have an influence on the outcome $Y$ that is not mediated by $T$; ii) $Z$ is not independence of the treatment $T$. 
\end{definition}

\begin{definition}[\textbf{Conditional Independence}~\cite{Pearl2000causality}]
Let $\textbf{X}=\lbrace x_1,...,x_n\rbrace$ be a finite set of variables. Let $P(\cdot)$ be a joint probability function 
over the variables in $\textbf{X}$, and let $\textbf{Y, W, Z}$ stand for any three subsets of variables in $\textbf{X}$. The 
sets $\textbf{Y}$ and $\textbf{W}$ are said to be conditionally independent given $\textbf{Z}$ if
\begin{equation} \label{eq:ci}
    P(y \mid w, z) = P(y\mid z), \mathrm{whenever} P(w, z)>0.
\end{equation}
In words, learning the value of $W$ does not provide additional information about $Y$, once we know $Z$, represented as $\textbf{Y} \indep \textbf{W} \mid \textbf{Z}$. 
\end{definition}

One of the most widely applicable tools for conditional independence is Kernel-based Conditional Independence test (KCI-test), whose test statistic is calculated from the kernel matrices, characterizing the uncorrelatedness of functions in reproducing kernel Hilbert spaces~\cite{zhang2011kernel}. 

\begin{definition}[\textbf{Back-Door}~\cite{Pearl2000causality}]
A set of variables $\textbf{Z}$ satisfies the back-door criterion relative to an ordered pair of variables $(x_i, x_j)$ in a Directed Acyclic Graph (DAG) if: 
(i) no node in $\textbf{Z}$ is a descendant of $x_i$; and 
(ii) $\textbf{Z}$ blocks every path between $x_i$ and $x_j$ that contains an arrow into $x_i$. 
Similarly, if $\textbf{Y}$ and $\textbf{W}$ are two disjoint subsets of nodes in the DAG, then $\textbf{Z}$ is said to satisfy the back-door criterion relative to $(\textbf{Y}, \textbf{W})$ if it satisfies the criterion relative to any pair $(x_i , x_j)$ such that $x_i \in \textbf{Y}$ and $x_j \in \textbf{W}$.
\end{definition}

\begin{definition}[\textbf{Front-Door}~\cite{Pearl2000causality}]
A set of variables $\textbf{Z}$ is said to satisfy the front-door criterion relative to an ordered pair 
of variables $(x_i , x_j)$ if: 
(i) $\textbf{Z}$ intercepts all directed paths from $x_i$ to $x_j$; 
(ii) there is no back-door path from $x_i$ to $\textbf{Z}$; and 
(iii) all back-door paths from $\textbf{Z}$ to $x_j$ are blocked by $x_i$.
\end{definition}

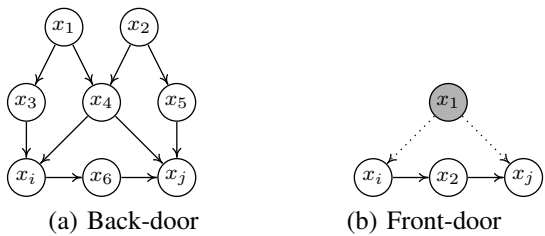
\begin{figure}[htp!] 
	\begin{center}
	\begin{tikzpicture}[scale=1.0, line width=0.5pt, inner sep=0.2mm, shorten >=.1pt, shorten <=.1pt]
		\draw (0, 0) node(xi) [circle, minimum size=0.5cm, draw] {{\footnotesize\,$x_i$\,}};
		\draw (1, 0) node(x6) [circle, minimum size=0.5cm, draw] {{\footnotesize\,$x_6$\,}};
		\draw (2, 0) node(xj) [circle, minimum size=0.5cm, draw] {{\footnotesize\,$x_j$\,}};
		
		\draw (0, 1) node(x3) [circle, minimum size=0.5cm, draw] {{\footnotesize\,$x_3$\,}};
		\draw (1, 1) node(x4) [circle, minimum size=0.5cm, draw] {{\footnotesize\,$x_4$\,}};
		\draw (2, 1) node(x5) [circle, minimum size=0.5cm, draw] {{\footnotesize\,$x_5$\,}};
		\draw (0.5, 2) node(x1) [circle, minimum size=0.5cm, draw] {{\footnotesize\,$x_1$\,}};
		\draw (1.5, 2) node(x2) [circle, minimum size=0.5cm, draw] {{\footnotesize\,$x_2$\,}};
		\draw[-arcsq] (xi) -- (x6) node[pos=0.5,sloped,above] {};
		\draw[-arcsq] (x6) -- (xj) node[pos=0.5,sloped,above] {};
		\draw[-arcsq] (x4) -- (xi) node[pos=0.5,sloped,above] {};
		\draw[-arcsq] (x4) -- (xj) node[pos=0.5,sloped,above] {}; 
		\draw[-arcsq] (x3) -- (xi) node[pos=0.5,sloped,above] {};
		\draw[-arcsq] (x1) -- (x4) node[pos=0.5,sloped,above] {};
		\draw[-arcsq] (x1) -- (x3) node[pos=0.5,sloped,above] {};
		\draw[-arcsq] (x2) -- (x4) node[pos=0.5,sloped,above] {};
		\draw[-arcsq] (x2) -- (x5) node[pos=0.5,sloped,above] {};
		\draw[-arcsq] (x5) -- (xj) node[pos=0.5,sloped,above] {};
		\end{tikzpicture}~~~~~~~~~~~~~~~~
	    \begin{tikzpicture}[scale=1.0, line width=0.5pt, inner sep=0.2mm, shorten >=.1pt, shorten <=.1pt]
		
		\draw (0, 1) node(xi) [circle, minimum size=0.5cm, draw] {{\footnotesize\,$x_i$\,}};
		\draw (1, 1) node(x2) [circle, minimum size=0.5cm, draw] {{\footnotesize\,$x_2$\,}};
		\draw (2, 1) node(xj) [circle, minimum size=0.5cm, draw] {{\footnotesize\,$x_j$\,}};
		\draw (1, 2) node(x1) [circle, fill=gray!60, minimum size=0.5cm, draw] {{\footnotesize\,$x_1$\,}};

		\draw[-arcsq] (xi) -- (x2) node[pos=0.5,sloped,above] {};
		\draw[-arcsq] (x2) -- (xj) node[pos=0.5,sloped,above] {};
		\draw[-arcsq,dotted] (x1) -- (xi) node[pos=0.5,sloped,above] {};
		\draw[-arcsq,dotted] (x1) -- (xj) node[pos=0.5,sloped,above] {}; 
		\end{tikzpicture}\\
		(a) Back-door ~~~~~~~~~~~~~~~(b) Front-door
		\caption{Illustration examples for back-door and front-door criteria, where unshaded variables are observed while the shaded one is unobserved. $x_1$ in (b) is a latent confounder. } 
	\label{fig:front}
	\end{center}
\end{figure}
The back-door and front-door criteria are two simple graphical tests to tell whether a set of variables $\textbf{Z} \subseteq \textbf{X}$ is sufficient to estimate the causal effect $P(x_j\mid {x_i})$. As illustrated in Fig.\ref{fig:front}, the set of variables $\textbf{Z} = \lbrace x_3, x_4\rbrace$ satisfies the back-door criterion, while $\textbf{Z} = \lbrace x_2\rbrace$ satisfies the front-door criterion.

The ladder of causality~\cite{pearl2018book} has gained much attention from many researchers of various domains. It states there are three types of levels for causation, i.e., association, intervention and counterfactuals. They correspond to different interactions with the data generation process:
\begin{definition}[\textbf{Association} (seeing)~\cite{Pearl2000causality}]
Let $\textbf{Y}$ and $\textbf{Z}$ be two subsets of variables in $\textbf{X}$. The association sentence "what the value of $\textbf{Z}$ is, if $\textbf{Y}$ is found as $y$?" is interpreted as the 
causal effects of $\textbf{Z}$ to $\textbf{Y}$, denoted by $P(\textbf{Z} \mid \textbf{Y})$.
\end{definition}
\begin{definition}[\textbf{Intervention} (doing)~\cite{Pearl2000causality}]
Let $\textbf{Y}$ and $\textbf{Z}$ be two subsets of variables in $\textbf{X}$. The intervention sentence "what the value of $\textbf{Z}$ would obtain, if $\textbf{Y}$ is assigned as $y$?" is interpreted as the causal effects of $\textbf{Z}$ to an action $do(\textbf{Y}=y)$, denoted by $P(\textbf{Z} \mid do(\textbf{Y}=y))$.
\end{definition}
\begin{definition}[\textbf{Counterfactuals} (imagining)~\cite{Pearl2000causality}]
Let $\textbf{Y}$ and $\textbf{Z}$ be two subsets of variables in $\textbf{X}$. The counterfactual sentence "what the value of $\textbf{Z}$ would have obtained, had $\textbf{Y}$ been $y$?" is interpreted as the potential response of $\textbf{Z}$ to an action $do(\textbf{Y}=y)$, which is the solution of $\textbf{Z}$ with the set of equations $f_{\textbf{Z}}(\cdot)$, denoted by $P(\textbf{Z}_{\textbf{Y}} \mid y)$.
\end{definition}

Assumptions~\ref{ass:cd1}-\ref{ass:cd3} below are usually made for causal discovery to find the causal structure: 
\begin{assumption}[\textbf{Causal Markov Assumption}~\cite{Pearl2000causality}]\label{ass:cd1}
A necessary and sufficient condition for a probability population distribution $P$ to be Markov relative to a causal graph (DAG) is that every variable is independent of all its nondescendants conditional on it parents.
\end{assumption}
\begin{assumption}[\textbf{Causal Faithfulness Assumption}~\cite{Pearl2000causality}]
\label{ass:cd2}
The probability distribution $P$ in the population has no additional conditional independence relations that are not entailed by d-separation\footnote{A set $\textbf{W}$ is said to d-separate $\textbf{Y}$ and $\textbf{Z}$ if and only if $\textbf{W}$ blocks every path from a variable in $\textbf{Y}$ to a variable in $\textbf{Z}$~\cite{Pearl2000causality}.} to the causal graph.
\end{assumption}
\begin{assumption}[\textbf{Causal Sufficiency Assumption}~\cite{spirtes2000causation}]\label{ass:cd3}
As for a set of variables $\textbf{X}$, there is no hidden common cause, i.e., latent confounder, that causes more than one variable in $\textbf{X}$.
\end{assumption}

Assumptions~\ref{ass:ci1}-\ref{ass:ci3} are commonly derived for causal inference to estimate the treatment effect: 
\begin{assumption}{\textit{(\textbf{Stable Unit Treatment Value Assumption}~\cite{sekhon2008neyman}):}} \label{ass:ci1}
The potential outcomes of any given unit do not vary with the treatments assigned to other units, and for each unit, there are no different versions of treatment, which lead to different potential outcomes.
\end{assumption}
\begin{assumption}[\textbf{Ignorability}~\cite{sekhon2008neyman}]
\label{ass:ci2}
Given the background covariates $X$, treatment assignment $T$ is independent to the potential outcomes, i.e., $T \indep Y(T=0), Y(T=1)\mid X$.
\end{assumption}
\begin{assumption}[\textbf{Positivity}~\cite{yao2021survey}]\label{ass:ci3}
Given any value of $X$, the treatment assignment $T$ is not deterministic:
\begin{equation}
    P(T=t \mid X=x)>0, \quad \forall \ t \ \mathrm{and} \ x.
\end{equation}
\end{assumption}

\begin{itemize}
    \item {\textit{Causal discovery}}
\end{itemize}
As for identifying causal structures from the data, a traditional way is to use interventions, randomized or control experiments, which is in many situations too expensive, too time-consuming, or even too unethical to conduct~\cite{peters2017elements}. Hence, uncovering causal information from purely observational data, known as causal discovery, has attracted much attention~\cite{glymour2019review,huang2018generalized,shimizu2006linear}.  
There are roughly two types of classic causal discovery approaches: constraint-based approaches and score-based ones. In the early of 1990's, constraint-based approaches exploited conditional independence relationships to recover the underlying causal structure among the observed variables, under appropriate assumptions. Such approaches include PC~\cite{spirtes2013causal}, and Fast Causal Inference (FCI)~\cite{spirtes1991algorithm,Pearl2000causality}, which allow different types of data distributions and causal relations, and could give asymptotically correct results. 
PC algorithm assumes that there is no latent confounder in the underlying causal graph; while FCI enables to handle the cases with latent confounders. %
However, what they recover belongs to an equivalence class of causal structures, which contains multiple DAGs entailing the same conditional independence relationships. On the other hand, the score-based approaches attempt to search for the equivalence class via optimizing a properly defined score function~\cite{chickering2002optimal,heckerman1995learning}, such as the Bayesian Information Criterion (BIC)~\cite{schwarz1978estimating}, the generalized score functions~\cite{huang2018generalized}, etc. 
And they output one or multiple candidate causal graphs with the highest score.
A well-known two-phase search procedure is Greedy Equivalence Search (GES) that searches directly over the space of equivalence classes~\cite{chickering2002optimal,glymour2019review}.  

To distinguish different DAGs in the equivalence class and enjoy the unique identifiability of the causal structure, algorithms based on the constrained functional causal models have emerged~\cite{huang2020causal}. These algorithms assumed the data generation mechanism, including the model class or noises' distribution: the effect variable is a function of direct causes along with an independent noise, as illustrated in Eq.\eqref{eq:scm} where causes $pa_i$ are independent with the noise $u_i$. This gives rise to the causal structure's unique identifiability, in that the model assumptions, e.g., the independence between $pa_i$ and $u_i$, only hold for the true causal direction but are violated for the wrong direction.     
Examples of these constrained functional causal models are linear non-Gaussian acyclic models (LiNGAM,~\cite{shimizu2006linear}), additive noise models (ANM,~\cite{hoyer2008nonlinear}), post-nonlinear models (PNL,~\cite{zhang2009identifiability}), etc. 

Furthermore, it was noted that there existed many significant but challenging topics that interest researchers. For instance, one may be interested in algorithms for the time series data. These algorithms include tsFCI~\cite{entner2010causal}, SVAR-FCI~\cite{malinsky2018causal}, tsLiNGAM~\cite{hyvarinen2008causal}, LPCMCI~\cite{gerhardus2020high}, etc.
Especially, Granger causality allows to infer causal structure of time series, with no instantaneous effects or latent confounders\footnote{Note that apart from tsLiNGAM and Granger causality, other mentioned methods from time series allow the presence of latent confounders.}. It has been previously and commonly applied in prediction of economics~\cite{granger1969investigating}. 
Constraint-based causal Discovery from heterogeneous/NOnstationary Data (CD-NOD) works on the cases where the underlying generation process changes across domains or over time~\cite{huang2020causal}. It uncovers the causal skeleton and directions, and estimates a low-dimensional representation for the changing causal modules.
%

\begin{itemize}
    \item {\textit{Causal inference}}
\end{itemize}
As for learning causal effects from the data, the most effective way is also to perform randomized experiments to compare the difference between the control and treatment groups. However, due to the high cost, practical and ethical issues, its applications are largely limited.  
Hence, estimating the treatment effect from observational data has gained increasing interests~\cite{rubin1974estimating,yao2021survey}.  

Difficulties of causal inference from observational data lie in the presence of confounder variables, which leads to i) the selection bias between the treated and control groups, and ii) spurious effects.
These problems would deteriorate the estimation performances of treatment outcomes.
To handle the problem of spurious effects, a representative method is the stratification, also known as subclassification or blocking~\cite{imbens2015causal}. The idea is to split the entire group into homogeneous subgroups, each of which in the treated and control groups shares similar characteristics over certain covariates~\cite{yao2021survey}. 
To conquer the selection bias challenge, there are generally two types of causal inference approaches~\cite{yao2021survey}. The first one aims at creating a pseudo group which is approximately consistent to the treated group. Such approaches include sample re-weighting methods~\cite{rosenbaum1983central}, matching methods~\cite{caliendo2008some}, tree-based methods~\cite{hill2011bayesian}, and representation based methods~\cite{johansson2016learning}, etc. 
The other type of approaches such as meta-learning based ones, first train the outcome estimation model on the observed data and then correct the estimation bias stemmed from the selection bias~\cite{kunzel2019metalearners,nie2021quasi}.

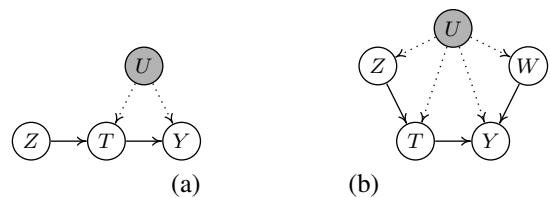
\begin{figure}[htp!] 
	\begin{center}
	\begin{tikzpicture}[scale=1.0, line width=0.5pt, inner sep=0.2mm, shorten >=.1pt, shorten <=.1pt]
		\draw (0, 0) node(Z) [circle, minimum size=0.5cm, draw] {{\footnotesize\,$Z$\,}};
		\draw (1, 0) node(T) [circle, minimum size=0.5cm, draw] {{\footnotesize\,$T$\,}};
		\draw (2, 0) node(Y) [circle, minimum size=0.5cm, draw] {{\footnotesize\,$Y$\,}};
		\draw (1.5, 1) node(U) [circle, fill=gray!60, minimum size=0.5cm, draw] {{\footnotesize\,$U$\,}};
		
		\draw[-arcsq] (Z) -- (T) node[pos=0.5,sloped,above] {};
		\draw[-arcsq] (T) -- (Y) node[pos=0.5,sloped,above] {};
		\draw[-arcsq,dotted] (U) -- (T) node[pos=0.5,sloped,above] {};
		\draw[-arcsq,dotted] (U) -- (Y) node[pos=0.5,sloped,above] {}; 

		\end{tikzpicture}~~~~~~~~~~~~~~~~
	    \begin{tikzpicture}[scale=1.0, line width=0.5pt, inner sep=0.2mm, shorten >=.1pt, shorten <=.1pt]
		
		\draw (0.5, 1) node(Z) [circle, minimum size=0.5cm, draw] {{\footnotesize\,$Z$\,}};
		\draw (2.5, 1) node(W) [circle, minimum size=0.5cm, draw] {{\footnotesize\,$W$\,}};
		\draw (1, 0) node(T) [circle, minimum size=0.5cm, draw] {{\footnotesize\,$T$\,}};
		\draw (2, 0) node(Y) [circle, minimum size=0.5cm, draw] {{\footnotesize\,$Y$\,}};
		\draw (1.5, 1.5) node(U) [circle, fill=gray!60, minimum size=0.5cm, draw] {{\footnotesize\,$U$\,}};
		
		\draw[-arcsq] (Z) -- (T) node[pos=0.5,sloped,above] {};
		\draw[-arcsq] (T) -- (Y) node[pos=0.5,sloped,above] {};
		\draw[-arcsq] (W) -- (Y) node[pos=0.5,sloped,above] {};
		
		\draw[-arcsq,dotted] (U) -- (T) node[pos=0.5,sloped,above] {};
		\draw[-arcsq,dotted] (U) -- (Y) node[pos=0.5,sloped,above] {};
		\draw[-arcsq,dotted] (U) -- (Z) node[pos=0.5,sloped,above] {};
		\draw[-arcsq,dotted] (U) -- (W) node[pos=0.5,sloped,above] {};
		\end{tikzpicture}\\
		(a) ~~~~~~~~~~~~~~~(b)
		\vspace{-0.2cm}
		\caption{Illustration examples for the instrumental variable and proxy variables, where unshaded variables are observed while the shaded one $U$ is unobserved. $T$ stands for the treatment, while $Y$ is the outcome. 
		In (a), $Z$ is an instrumental variable while $\lbrace Z, W \rbrace$ are proxy variables in (b).} 
	\label{fig:proxy}
	\end{center}
\end{figure}

The above-mentioned causal inference methods rely on the satisfactions of Assumption~\ref{ass:ci1}-\ref{ass:ci3}. In practice, such assumptions may not always hold. For instance, when latent confounders exist, Assumption~\ref{ass:ci2} does not hold, i.e., $T \not\!\perp\!\!\!\perp Y(T=0), Y(T=1)\mid X$. In such a case, one solution is to apply sensitivity analysis to investigate how inferences might change with various magnitudes of given unmeasured confounders. Sensitivity analysis quantifies the unmeasured confounding or hidden bias generally via differences between the unidentifiable distribution $P(Y(T=t)\mid T=1-t, X)$ and the identifiable distribution $P(Y(T=t)\mid T=t, X)$,
\begin{equation}
\begin{aligned}
    c_t(X) &= \mathbb{E}(Y(T=t)\mid T=1-t, X) \\ 
    &- \mathbb{E}(Y(T=t)\mid T=t, X).
\end{aligned}
\end{equation}
Specifying bounds for $c_t(X)$, one could obtain the bounds for the expectation of outcomes $\mathbb{E}(Y(T=t))$, in the form of unidentifiable selection bias~\cite{robins2000sensitivity,tan2006distributional}.
Another possible solutions are to make the best of Instrumental Variables (IV) regression methods and Proximal Causal Learning (PCL) methods. These methods are used to predict the causal effects of the treatment or the policy, even with the existence of latent confounders~\cite{chen2021instrumental,li2021causal,xu2021deep,miao2018identifying}.
%
It is worthwhile to note that the intuition behind PCL is to construct two conditionally independent proxy variables, so as to reflect the the effects of unobserved confounders~\cite{miao2018identifying,xu2021deep}.
Figure~\ref{fig:proxy} illustrates examples of the IV and proxy variables.

\subsection{Reinforcement Learning}
In this section, we first offer RL's characteristics and basic concepts, following the standard textbook definitions~\cite{sutton1998introduction}. And thereafter we give baselines from perspectives of model-free and model-based RL methods, which implies that the world model (including dynamics and reward functions) is utilized or not.

\begin{itemize}
    \item {\textit{Definitions}}
\end{itemize}
Compared with supervised and unsupervised learning, RL enjoys advantages from two key components: optimal control, and trial and error.
Based on the optimal control problem, Richard Bellman developed a dynamic programming approach, using the value function with the system's state information to give mathematical formalization~\cite{bellman1966dynamic,nian2020review}. Such a value function is known as the Bellman equation, given as, 
\begin{equation}
    V(s_t) = r(s_t) + \gamma \sum_{s_{t+1} \in \mathcal{S}} P(s_{t+1}\mid s_t,a_t) \cdot V(s_{t+1}),
\end{equation}
where $V(s_t)$ is the value function of the state $s_t$ at time $t$, $s_{t+1}$ is the next state, $r(s_t)$ is the reward function, and $\gamma$ is a discount factor. $\mathcal{P}(s_{t+1}| s_t,a_t)$ is the transition probability of $s_{t+1}$ given the current state $s_t$ and action $a_t$.
Learning through interactions is the essence of RL. An agent interacts with the environment by taking an action $a_t$ in state $s_t$, and once observing its next state $s_{t+1}$ and the reward $r(s_t)$, it needs to adjust the policy to strive for the optimal returns~\cite{sutton2018reinforcement,henderson2018deep,mnih2015human}. Such a trial-and-error learning regime is originated from animal psychology, which implies that actions accounting for good outcomes are likely to be repeated, which those for bad outcomes are muted~\cite{nian2020review,thorndike1898animal}.

RL addresses the problem of learning a policy from the available information in different settings, including Multi-Armed Bandits (MAB), Contextual Bandits (CB), Markov Decision Process (MDP), Partially Observed Markov Decision Process (POMDP), Imitation Learning (IL), and Dynamic Treatment Regime (DTR).
\begin{definition}[\textbf{Markov Decision Process (MDP)}~\cite{sutton1998introduction}]
A Markov decision process is defined as a tuple $\mathcal{M}=(\mathcal{S},\mathcal{A}, \mathcal{P},\mathcal{R},\gamma)$, where $\mathcal{S}$ is a set of states $s\in \mathcal{S}$ and $\mathcal{A}$ is a set of actions $a \in \mathcal{A}$. $\mathcal{P}$ denotes the state transition probability that defines the distribution $\mathcal{P}(s_{t+1}|s_t,a_t)$, while $\mathcal{R}:\mathcal{S}\times \mathcal{A} \to \mathbb{R}$ denotes the reward function. $\gamma \in \lbrack 0,1\rbrack$ is a discount factor.
\end{definition}
\begin{definition} 
{\textit{(\textbf{Partially Observed Markov Decision Process (POMDP)}~\cite{sutton1998introduction}):}}
\label{def:pomdp}
The partially observed Markov decision process is defined as a tuple $\mathcal{M}=(\mathcal{S},\mathcal{A}, \mathcal{O},\mathcal{P},\mathcal{R},\mathcal{E},\gamma)$, where $\mathcal{S},\mathcal{A}, \mathcal{P}, \mathcal{R}, \gamma$ are the same notations as MDP. $\mathcal{O}$ denotes the set of observations $o \in \mathcal{O}$ while $\mathcal{E}$ is an emission function that determines the distribution $\mathcal{E}(o_t\mid s_t)$.
\end{definition}
%
\begin{definition}[\textbf{Multi-Armed Bandits (MAB)}~\cite{bareinboim2015bandits}]
A K-armed bandit problem is defined as a tuple $\mathcal{M}=(\mathcal{A}, \mathcal{R})$, where $\mathcal{A}$ is a set of player's arm choices from $K$ arms $a_t \in \mathcal{A}=\lbrace a_1,...,a_K\rbrace$ at round $t$, and $\mathcal{R}$ is a set of outcome variables representing the rewards $r_t \in \lbrace 0,1\rbrace$. 
\end{definition}

Note that when unobserved confounders exist in the K-armed bandit, the model then would be established and replaced by $\mathcal{M}=(\mathcal{A}, \mathcal{R},U)$, where $U$ is the unobserved variable that implies the payout rate of arm $a_t$ and the propensity score for choosing arm $a_t$~\cite{bareinboim2015bandits}.
%
\begin{definition}[\textbf{Contextual Bandits (CB)}~\cite{langford2007epoch}]
The contextual bandit is defined as a tuple $\mathcal{M}=(\mathcal{X},\mathcal{A}, \mathcal{R})$, where $\mathcal{A}$ and $\mathcal{R}$ are the same notations as MAB. $\mathcal{X}$ is a set of contexts, i.e., the observed side information. 
\end{definition}
%
\begin{definition}[\textbf{Imitation Learning Model (IL)}~\cite{samsami2021causal}]
An imitation learning model is defined as a tuple $\mathcal{M}=(\mathcal{O}, \mathcal{T})$, where $\mathcal{O}$ denotes the accessible high-dimensional observations $o \in \mathcal{O}$, while $\mathcal{T}$ denotes the trajectories generated by the expert policy $\mathcal{T} \sim \pi_D (\cdot \mid o)$. 
\end{definition}
\begin{definition}[\textbf{Dynamic Treatment Regime (DTR)}~\cite{murphy2003optimal,zhang2020designing}] \label{def:DTR}
A dynamic treatment regime is defined as a sequence of decision rules $\lbrace \pi_T:\forall T \in \textbf{T} \rbrace$, where $\textbf{T}$ is a set of treatments. Each $\pi_T$ is a mapping function from the values of treatments' and covariates' history $H_T$ to the domain of probability distributions over $T$, denoted by $\pi_T(T\mid H_T)$.
\end{definition}

\begin{itemize}
    \item {\textit{Model-free methods}}
\end{itemize}
Model-free RL methods typically have irreversible access to the world model, but learn the policy directly and purely from interactions with the environment, similar to the way we act in the real world~\cite{moerland2020model}. Popular approaches include policy-based, value-based and actor-critic ones.

Policy-based methods directly learn the optimal policy $\pi^*$ via the policy parameter $\theta$ to maximize the accumulated reward. They basically take the best of policy gradient theorem for deriving $\theta$. Typical methods are Trust Region Policy Optimization (TRPO)~\cite{schulman2015trust}, Proximal Policy Optimization (PPO)~\cite{schulman2017proximal}, etc., which use function approximation to adaptively or artificially adjust super-parameters to accelerate the methods' convergence. 


In value-based methods, the agent updates the value function to gain the optimal one $Q^*(s,a)$, implicitly obtaining a policy. Q-learning, state–action–reward–state–action (sarsa) and Deep Q-learning Network (DQN) are typical value-based methods~\cite{sutton1998introduction,mnih2013playing,mnih2015human,sutton2018reinforcement}. The update rules for Q-learning and sarsa involve a learning rate $\alpha$ and a temporal difference error $\delta_t$:
\begin{equation}
    Q(s_t,a_t) = Q(s_t,a_t) + \alpha \delta_t,
\end{equation}
where $\delta_t = r_{t+1} + \gamma \max_{a_{t+1}} Q(s_{t+1},a_{t+1}) - Q(s_{t},a_{t})$ in off-policy Q-learning, and $\delta_t = r_{t+1} + \gamma Q(s_{t+1},a_{t+1}) - Q(s_{t},a_{t})$ in on-policy sarsa. However, it can only handle discrete spaces of states and actions. DQN with deep learning expresses the value or policy with a neural network, enabling to deal with continuous states or actions. It stabilizes the Q-function learning by experience replay and the frozen target network~\cite{mnih2015human}. Improvements of DQN are Double DQN~\cite{van2016deep}, Dueling DQN~\cite{wang2016dueling}, etc.

Actor-critic methods combine the merits of networks from policy-based and value-based methods, where the actor network originates from the policy-based methods while the critic network from the value-based ones~\cite{konda1999actor,mnih2016asynchronous}. The principle framework of actor-critics consists of two parts: i) actor: outputting the best action $a_t$ based on a state $s_t$, which controls how the agent behaves with learning the optimal policy; ii) critic: computing the Q value of the action, which achieves evaluation of the policy.
Typical methods include Advantage actor-critic (A2C)~\cite{mnih2016asynchronous}, Asynchronous Advantage actor-critic (A3C)~\cite{mnih2016asynchronous}, Soft actor-critic (SAC)~\cite{haarnoja2018soft}, Deep Deterministic Policy Gradient (DDPG)~\cite{lillicrap2015continuous}, etc. Especially, SAC introduces a maximum entropy term to improve the agent's exploration and stability of the training process with the stochastic policy~\cite{haarnoja2018soft}; DDPG applies neural networks to operate on high-dimensional and visual state space. It embraces Deterministic Policy Gradient (DPG)~~\cite{silver2014deterministic} and DQN~\cite{mnih2013playing} methods to be roles of the actor and critic, respectively, alleviating the high bias and high variance problems~\cite{lillicrap2015continuous}.

\begin{itemize}
    \item {\textit{Model-based methods}}
\end{itemize}
Without interacting with the environment directly, model-based RL methods majorly exploit the learned or given world model to simulate transitions, optimizing the target policy efficiently. It is similar to the way humans imagine in their mind~\cite{moerland2020model}. We here introduce some common model-based RL algorithms according to how the model is used, i.e., the black-box model for trajectory sampling and white-box model for gradient propagation~\cite{luo2022survey}.

With an available black-box model, the direct way of applying it in policy learning is to plan in such a model. Methods include Monte Carlo (MC)~\cite{nagabandi2018neural}, Probabilistic Ensembles with Trajectory Sampling (PETS)~\cite{chua2018deep}, Monte Carlo Tree Search (MCTS)~\cite{chaslot2008monte,silver2016mastering,silver2017mastering}, etc. MCTS is an extension sampling method of MC, and it basically adopts a tree search to determine the action with high probability to transit to the high-value state at each time step~\cite{chaslot2008monte}. It has been applied in AlphaGo and AlphaGo Zero, to challenge the professional human players in the game of Go~\cite{silver2016mastering,silver2017mastering}.
On the other hand, an available model could be used to generate simulated samples to accelerate the policy learning or value approximation, and this process is well-known as the Dyna-style methods~\cite{sutton1990integrated}. That is, models in the Dyna play a role of experience generator to produce augmented data. For instance, Model Ensemble Trust Region Policy Optimization (ME-TRPO)~\cite{kurutach2018model} learns a set of dynamic models with the collected data, with which it generates imagined experiences. Thereafter it updates the policy with augmented data in the ensemble model, using TRPO model-free algorithm~\cite{schulman2015trust}; Model-Based Policy Optimization (MBPO)~\cite{janner2019trust} samples the branched rollouts with the policy and the learned model, and utilizes SAC~\cite{haarnoja2018soft} to further learn the optimal policy with augmented data.

With a white-box dynamics model, one can employ the internal structure to generate gradients of the dynamics so as to facilitate the policy learning. Typical methods include Guided Policy Search (GPS)~\cite{levine2013guided}, Probabilistic Inference for Learning Control (PILCO)~\cite{deisenroth2011pilco}, etc. GPS~\cite{levine2013guided} utilizes the path optimization technique to guide the training process, resulting in improved efficiency. It draws samples via the Iterative Linear Quadratic Regulator (iLQR), which are used to initialize the neural network policy and further update the policy~\cite{luo2022survey}. PILCO~\cite{deisenroth2011pilco} assumes the dynamic model as the Gaussian process. After learning such a dynamic model, it evaluates the policy with approximate inference and acquires the policy gradients for policy improvement.   
In real-world applications, offline RL often matters, where the agent has to learn a satisfactory policy only from an offline experience dataset but with no interaction with the environment. One key challenge of offline RL belongs to the distributional shift problem, resulting from the discrepancy between the behavior policy for the training data and the current learning policy~\cite{luo2022survey}. To bypass the distributional shift problem,~\cite{yu2020mopo} proposed a Model-Based Offline Policy Optimization (MOPO) algorithm. They derive a policy value lower bound based on the learned model and attempt to penalize the rewards by the uncertainty of the dynamics.     

\section{Causal Reinforcement Learning}\label{sec:crl}
Since causality and reinforcement learning can be umbilically tied with each other, it is desirable to investigate how they are effectively integrated to implement policy learning tasks or(and) causality tasks~\footnote{Please see Appendix for preliminary of causality and RL.}. And such an implementation is called causal reinforcement learning, which is defined formally as follows.
\begin{definition}[\underline{C}ausal \underline{R}einforcement \underline{L}earning, CRL] CRL is a suite of algorithms which aim at embedding causal knowledge into RL for more efficient and effective model learning, policy evaluation, or policy optimization. It is formalized as a tuple $(\mathcal{M}, \mathcal{G})$, where $\mathcal{M}$ represents an RL model setting, e.g., MDP, POMDP, MAB, etc., while $\mathcal{G}$ stands for the causal-based information regarding an environment or task, e.g., causal structure, causal representation or features, latent confounders, etc. 
\end{definition}

Based on whether the causal information is empirically given or not, methods of causal reinforcement learning roughly fall into two categories, i.e., (i) methods based on the given or assumed causal information; and (ii) methods based on the unknown causal information which has to be learned by techniques. 
Causal information includes especially the causal structure, and causal representation or causal features, latent confounders, etc.

\subsection{CRL with Prior Causal Information} \label{sec:known}
Here we review CRL methods where the causal information is known (or given a prior) explicitly or implicitly. Generally, such methods assume the causal structure regarding to the environment or task is given a prior from the experts, which may include how many latent factors are, where the latent confounders locate or how they affect other observed variables of interest. 
As for the latent confounders scenarios, most of them would consider to eliminate the confounding bias towards the policy via proper techniques, while they use RL methods to learn the optimal policy. They may also theoretically prove the worst-case bounds of policy to achieve policy evaluation. 
As for the unconfounded scenarios, they use prior causal knowledge for sample efficiency, causal explanation or generalization in policy learning. During or before the modern RL procedure, these methods perform data augmentation with causal mechanisms; narrow the search space with causal information; or give preference towards those having causal influence. 

We summarize these CRL methods according to different model settings, i.e., MDP, POMDP, bandits, DTR and IL\footnote{Please note that since there are close connections between these model settings, it is not restricted for a CRL method to strictly belong to one of these models only.}.

\begin{table}[ht]	\label{tab:summary_known}
	\caption{Part of CRL algorithms with known causal information}
	\begin{center} 
	\begin{tabular}{  m{1cm} m{7cm} } 
	    \toprule
		\multicolumn{1}{c}{\bf Models}  &\multicolumn{1}{c}{\bf Algorithms} \\ 
		\midrule
		MDP & IVOPE~\cite{chen2022instrumental}, IV-SGD and IV-Q-Learning~\cite{li2021causal}, IVVI~\cite{liao2021instrumental}, CausalDyna~\cite{zhu2021causaldyna}, CTRL$_g$ and CTRL$_p$~\cite{lu2020sample}, IAEM~\cite{zhu2022invariant}, MORMAXC~\cite{zhang2016markov}, DOVI~\cite{wang2021provably}, FQE~\cite{bruns2021model}, COPE~\cite{shi2022off}, off\_policy\_confounding \cite{namkoong2020off}, RIA~\cite{guo2022relational}, 
        etc.\\
		\midrule
		POMDP & CF-GPS~\cite{buesing2018woulda}, Gumbel-Max SCMs~\cite{oberst2019counterfactual}, CFPT~\cite{killian2022counterfactually}, Decoupled POMDPs~\cite{tennenholtz2020off}, PCI~\cite{bennett2021proximal}, partial history weighting~\cite{hu2022off}, COMA~\cite{foerster2018counterfactual}, influence MOA~\cite{jaques2019social}, CCM~\cite{barton2018measuring}, Confounded-POMDP~\cite{shi2022minimax}, etc.\\
		\midrule
		Bandits & Causal TS~\cite{bareinboim2015bandits}, DFPV~\cite{xu2021deep}, TS$^{RDC*}$~\cite{forney2017counterfactual}, SRIS~\cite{sen2017identifying}, Causal Bandit~\cite{yabe2018causal,lattimore2016causal}, 
        C-UCB and C-TS~\cite{lu2020regret}, UCB+IPSW~\cite{li2021unifying}, B-kl-UCB and B-TS~\cite{zhang2017transfer}, POMIS~\cite{lee2018where,lee2019structural}, Unc\_CCB~\cite{subramanian2022causal}, OptZ~\cite{bennett2019policy}, CRLogit~\cite{kallus2018confounding,kallus2021minimax}, etc.  \\
		\midrule
		DTR  & OFU-DTR and PS-DTR~\cite{zhang2020designing}, UC$^c$-DTR~\cite{zhang2019near}, CAUSAL-TS$^*$~\cite{zhang2022online}, IV-optimal and IV-improved DTRs~\cite{chen2021estimating}, etc.  \\
		\midrule
		IL  & CI~\cite{zhang2020causal}, exact linear transfer method~\cite{etesami2020causal}, Sequential $\pi$-Backdoor~\cite{kumor2021sequential}, CTS~\cite{tennenholtz2022covariate}, DoubIL and ResiduIL~\cite{swamy2022causal}, MDCE-IRL and MACE-IRL~\cite{zhou2017infinite}, etc.
		\\
		\bottomrule
	\end{tabular}
	\end{center}
\end{table}

\subsubsection{MDP}
\begin{figure}[t!]
    \centering
    \includegraphics[scale=0.35]{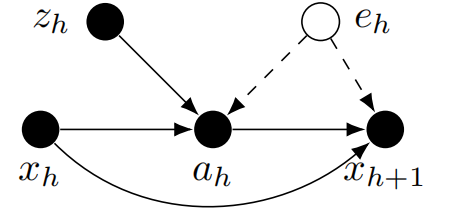}
    \caption{An example of causal graphical illustration of confounded MDP, where $\lbrace z_h \rbrace$ is a sequence of IVs, $\lbrace e_h \rbrace$ is a sequence of unobserved confounders, and $x_h$ is the current state variable at the $h^{th}$ time step~\cite{liao2021instrumental}.}
    \label{fig:liao2021instrumental}
\end{figure}

\begin{figure}[t!]
\centering
\subfigure[Offline setting]{
\includegraphics[width=0.22\textwidth]{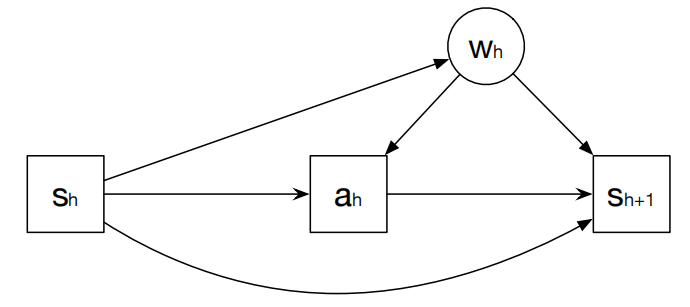}} 
\subfigure[Online setting]{
\includegraphics[width=0.22\textwidth]{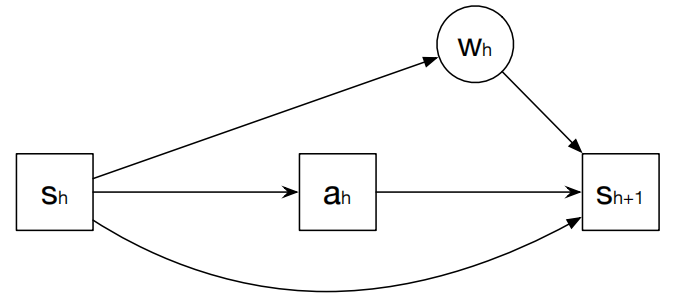}}
\caption{An example of causal graphical illustration of confounded MDP in the offline and online settings, where $\lbrace w_h \rbrace$ is a sequence of unobserved confounders at the $h^{th}$ time step~\cite{wang2021provably}.}
\label{fig:wang2021provably}
\end{figure}

Giving a digital advertising example, Li et al.,~\cite{li2021causal} showed the existence of reinforcement bias in policy, which could be amplifiable in interaction with the environment. To correct the bias theoretically and practically, they proposed a class of instrumental variable-based RL methods under the two-timescale stochastic approximation framework, incorporating a stochastic gradient descent routine, and Q-learning algorithm, etc. They considered the MDP environment where noises may be correlated with states or actions, based on which they constructed the IVs. IVs prescribed state-dependent decisions. They used instrumental variables (IVs) to learn the causal effect of a policy, based on the given causal structure, and hereafter true rewards were debiased and the optimal policy was learned.
%
Focusing on the case~\cite{kallus2020confounding} where the confounder $u_t$ at time $t$ affected both the action $a_t$ and the state $s_{t+1}$ as illustrated in Fig.~\ref{fig:liao2021instrumental}, Liao et al.,~\cite{liao2021instrumental} constructed a Confounded MDP model taking IVs and unobserved confounders (UCs) into consideration, namely CMDP-IV. 
They derived a conditional moment restriction, by which they identified the confounded nonlinear transition dynamics under the additive UCs assumption. 
With the primal-dual formalization of such a conditional moment restriction, they eventually proposed an IV-aided Value Iteration (IVVI) algorithm to learn the optimal policy in offline RL.
%
Zhang et al.,~\cite{zhang2016markov} investigated the problem of estimating the MDP with the presence of unobserved confounders. If ignoring such confounders, sub-optimal policies may be obtained. Hence, they utilized causal languages to formalize the problem, and demonstrated explicitly two classes of policies (i.e., the experimental policy and the counterfactual one). After proving the superiority of the counterfactual policy instead of the experimental policy, they restricted standard MDP methods to search in the space of counterfactual policies, with empowered performance in speed and convergence than state-of-the-arts algorithms.
%
Wang et al.,~\cite{wang2021provably} studied the problem of constructing the information gains from the offline dataset so as to improve the sample efficiency in the online setting. Hence, they proposed a Deconfounded Optimistic Value Iteration algorithm (DOVI). 
As shown in Fig.~\ref{fig:wang2021provably}, in the offline setting, they assumed the confounders were partially observed, whose confounding bias could be corrected by the backdoor criterion; while in the online setting, they assumed the confounders were unobserved, whose confounding bias could be corrected by intermediate states via the frontdoor adjustments. 
They eventually gave the regret analysis of their proposals.

To handle the issue of data scarcity and mechanism heterogeneity, Lu et al.,~\cite{lu2020sample} proposed a sample-efficient RL algorithm that leveraged SCMs to model the state dynamics process, and aimed for the general policy over the population as well as the personalized policy over each individual. In particular, as for the general policy, based on the causal structure in a MDP, they assumed the state variable $s_{t+1}$ satisfies the SCM,
\begin{equation}
    s_{t+1} = f(s_t, a_t, u_{t+1}),
\end{equation}
where $f$ is a function representing the causal mechanism from the causes to $s_{t+1}$, $a_t$ is the action at time $t$, and $u_{t+1}$ stands for the noise term of $s_{t+1}$.
As for the personalized policy for an individual, they assumed the state variable $s_{t+1}$ satisfies the SCM,
\begin{equation}
    s_{t+1} = f(s_t, a_t, \theta_c, u_{t+1}),
\end{equation}
where $f$ represents the overall family of causal mechanisms, and $\theta_c$ captures change factors that may vary cross individuals.
With these two state transition processes, they then employed Bidirectional Conditional generative adversarial network framework to estimate $f$, $u_{t+1}$ and $\theta_c$. With performing the counterfactual reasoning, the data scarcity problem was mitigated.
Zhu et al.,~\cite{zhu2021causaldyna} adopted the same idea of counterfactual reasoning based on SCMs to improve the sample efficiency. They modeled the time-invariant property (e.g., the mass of an object in robotic manipulations) as an observed confounder affecting states variables at all time steps, and proposed a Dyna-style causal RL algorithm.
%
Following the goal-conditioned MDP model~\cite{nair2019causal}, Feliciano et al., ~\cite{feliciano2021causal} accelerated the exploration and exploitation learning processes to learn policies via causal knowledge. Assuming the presence of a possibly incomplete or partially correct casual graph, they narrowed the search space of actions via graph queries, and developed a method to guide the action selection policy.
%
To efficiently perform RL tasks, Lu et al.,~\cite{lu2021efficient} studied causal MDPs, where interventions and states/rewards constituted a three-layer causal graph. The states or rewards (outcomes) was on the top layer, parents of the outcome were on the middle layer, while the direct interventions (manipulable) were on the bottom layer. Then given such prior causal knowledge, they proposed the causal upper confidence bound value iteration (C-UCBVI) algorithm and causal factored upper confidence bound (CF-UCBVI) algorithm, to circumvent the curse of dimensionality of actions or states. They finally proved the regret bounds for verification.

As for accelerating the explanation for decisions of the system, Madumal et al.,~\cite{madumal2020explainable} employed causal models to derive causal explanations of the behavior in model-free RL agents. 
Specifically, they introduced an \textit{action influence model} where the variables in the structural causal graph were not only states but also actions.
They assumed that a DAG with causal directions was given a prior, and presented an approach to learn such a structural causal model so as to generate contrastive explanations for why and why not questions. Finally, they were enable to explain why new events happened by counterfactual reasoning.
%
As for the data efficiency and generalization problems, Zhu et al.,~\cite{zhu2022invariant} introduced an invariance of action effects among states, which was inspired by the fact that the same action could have similar effects among different states on the transitions.
Utilizing such an invariance, they proposed a dynamics-based method termed as IAEM (Invariant Action Effect Model) for generalization. 
Specifically, they first characterized the invariance as the residual between representations from neighboring states. And they applied a contrastive-based loss and a self-adapted weighting strategy for better estimation of these invariance representations.
%
%
Guo et al.,~\cite{guo2022relational} proposed a relation intervention MBRL method to generalize to dynamics unseen environments, where a latent factor that changed across environments was introduced to characterize the shift of the transition dynamics. Such a factor was drawn from the history transition segments. Given the causal graph, they introduced interventional prediction module and a relational head to reduce the redundant information from the factors.

Off-Policy Evaluation (OPE) of sequential decision-making policies under uncertainty is a fundamental and essential problem in batch RL. However, there existed challenges which hammered its development, including the confounded data, etc.
%
The presence of confounders in data would render the evaluation of new policies unidentifiable. To handle such an issue, Kallus et al.,~\cite{kallus2020confounding} explored the partial identification for off-policy evaluation in the setting of infinite-horizon RL with confounders. Specifically, they first assumed the stationary distributions for unobserved confounding and the subjection to a sensitivity model for the data. In their satisfied causal model, the confounder $u_t$ at the timestep $t$ affected both the action $a_t$ and the state $s_{t+1}$.
They computed the sharpest bound of the policy value by characterizing and optimizing the partially identified set. They ultimately proved their proposed approximation method was consistent to establish the bounds in practice.
%
In the cases where the observed action $a_t$, state $s_{t+1}$ and reward $r_t$ were confounded by the i.i.d. observed variable $u_t$ at each time step $t$, Bennett et al.,~\cite{bennett2021off} proposed an OPE method in the infinite-horizon setting, to estimate the stationary distribution ratio of states in the presence of unobserved confounders and to estimate the policy value while avoiding direct modeling of reward functions. They proved the identifiability of the confounding model,
They also gave statistical consistency and provided error bounds under some assumptions.
%
When encountering unobserved confounding, Namkoong et al.,~\cite{namkoong2020off} analyzed the sensitivity of OPE methods in sequential decisions making cases, and demonstrated that even a small amount of confounding could induce heavily bias. Hence, they proposed a framework to quantify the impact of the unobserved confounding and to compute the worst-case bounds, which restricted the confounding in a single time step, i.e., the confounders might only directly affect one of the many decisions and further influence future rewards or actions. The bounds on the expected cumulative rewards were estimated by an efficient loss-minimization-based optimization, with the statistical consistency.
%
Both previous methods~\cite{kallus2020confounding,namkoong2020off} adopted importance sampling approaches to compute the worst-case bounds of a new policy. While Kallus et al.,~\cite{kallus2020confounding} constructed bounds in the infinite horizon case and Namkoong et al.,~\cite{namkoong2020off} deduced the worst-case bounds with restricted confounding at a single time step, they however could not render the finite case with confounding at each step tractable. To conquer this challenge, Bruns~\cite{bruns2021model} developed a model-based robust MDP approach to calculate finite horizon lower bounds with confounders that were drawn i.i.d. each period, combining the sensitivity analysis. For the cases where the confounders were persistently available, off-policy evaluation was far more challenging.
%
Recent methods for OPE in the MDP or POMDP models~\cite{namkoong2020off,kallus2020confounding,bennett2021off,tennenholtz2020off,nair2021spectral,shi2022minimax} did not considered the case which aimed at estimating confidence intervals offline for the target policy's value in infinite horizons. Shi et al.,~\cite{shi2022off} took great care of such a case. They modeled the data generation process based on a confounded MDP additionally with some observed immediate variables, and named this causal diagram as CMDPWM (Confounded MDP With Mediators). With the mediators, the target policy's value was proved to be identifiable. And they provided an algorithm to estimate the off-policy values robustly.
Their method helped ride-sharing companies solve the problem of evaluating different customer recommendation programs that may contained latent confounders.
%
If ignoring confounders when minimizing the Bellman error, one might obtain biased Q-function estimates. Chen et al.,~\cite{chen2021instrumental} exploited both IVs and RL in the language of OPE, proposing a class of IV methods to overcome these confounders and achieve identifying the value of a policy.
They conducted experiments based on the MDP environments with a set of OPE benchmark problems and various IV comparisons.
\begin{figure}[t]
    \centering
    \includegraphics[scale=0.25]{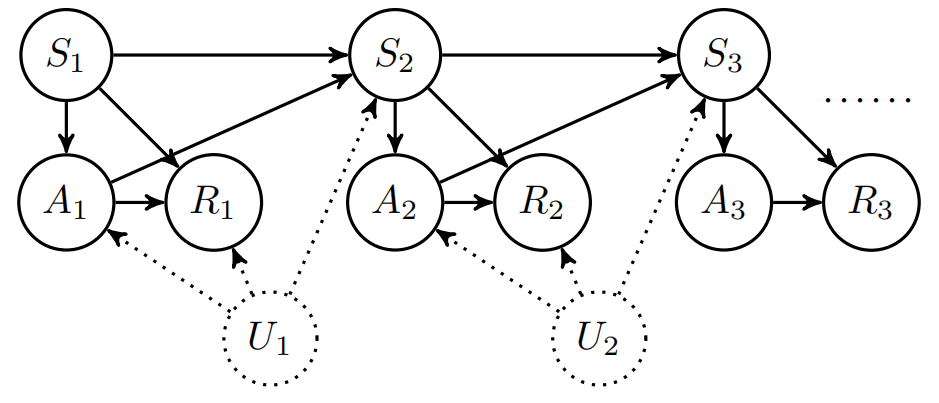}
    \caption{An example of causal graphical illustration of MDP with unmeasured confounders, where $U_{\cdot}$ are confounders. The action policy, state transition and reward emission are all confounded at every step~\cite{bennett2021off}.}
    \label{fig:bennett2021off}
\end{figure}

\begin{figure}[t]
    \centering
    \includegraphics[scale=0.5]{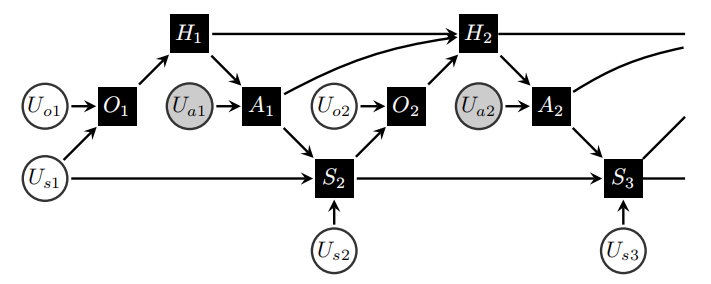}
    \caption{An example of causal graphical illustration of POMDP, where $U_{\cdot}$ are noise variables, and $H_t$ are histories. The mechanism that generates the actions $A_t$ relies on the histories $H_t$~\cite{buesing2018woulda}.}
    \label{fig:buesing2018woulda}
\end{figure}
\subsubsection{POMDP}
To shrink the mismatch between the model and the true environment as well as improve sample efficiency, Buesing et al.,~\cite{buesing2018woulda} leveraged counterfactual reasoning and proposed the Counterfactually-Guided Policy Search (CF-GPS) algorithm for learning policies in POMDPs from off-policy experience. 
They first modeled POMDPs using the SCM representation as demonstrated in Fig.~\ref{fig:buesing2018woulda}, and showed how to use the given structural causal models to counterfactually evaluate the arbitrary policies on individual off-policy episodes. 
Further, with available logged experience data, they made generalization to the policy search case via the counterfactual analysis in SCMs. Here they assumed the transition function is deterministic. 
%
To extend the counterfactual reasoning from deterministic transition functions to stochastic transitions, Oberst et al.,~\cite{oberst2019counterfactual} made further assumptions of a specific SCM, that is, the Gumbel-Max SCM, and proposed an off-policy evaluation method via counterfactual queries in categorical distributions. In particular, such a given class of SCM enabled to generate counterfactual trajectories in finite POMDPs. They identified those episodes where the counterfactual difference in reward is most dramatic, which could be used to flag individual trajectories for review by domain experts.
%
Given similarly as~\cite{oberst2019counterfactual} that the POMDP followed a particular Gumbel-Max SCM, Killian et al.,~\cite{killian2022counterfactually} presented a Counterfactually Guided Policy Transfer algorithm to fulfill the policy transfer from the source domain to a target domain offline, off-policy clinical settings. They followed the identical model formalization as~\cite{buesing2018woulda}, POMDP with SCMs. Intuitively, they leveraged the observed transition probability and the trained treatment policy in the source domain to aid guiding the counterfactual sampling in the target domain.
%
Instead of using counterfactual analysis, Tennenholtz et al.,~\cite{tennenholtz2020off} took the best of the effects of intervention on the world with a different action and an evaluating policy, and used them to fulfill the off-policy evaluation of sequential decisions. Their work was majorly based on the POMDPs, and especially, the Decoupled POMDPs where state variables were partitioned into two disjoint set, the observed and the unobserved ones. They showed that such decoupled new models could reduce the estimation bias from the general POMDPs. However, their method was only valid in tabular settings where states and actions were discrete.

\begin{figure}[t!]
\centering
\subfigure[Under logging policy]{
\includegraphics[width=0.28\textwidth]{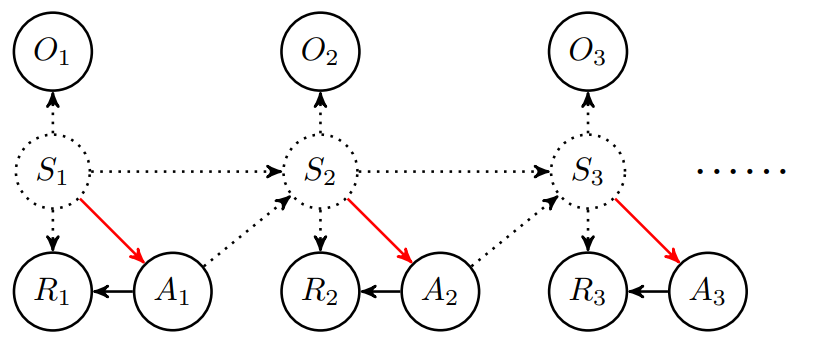}} 
\subfigure[Under evaluation policy]{
\includegraphics[width=0.35\textwidth]{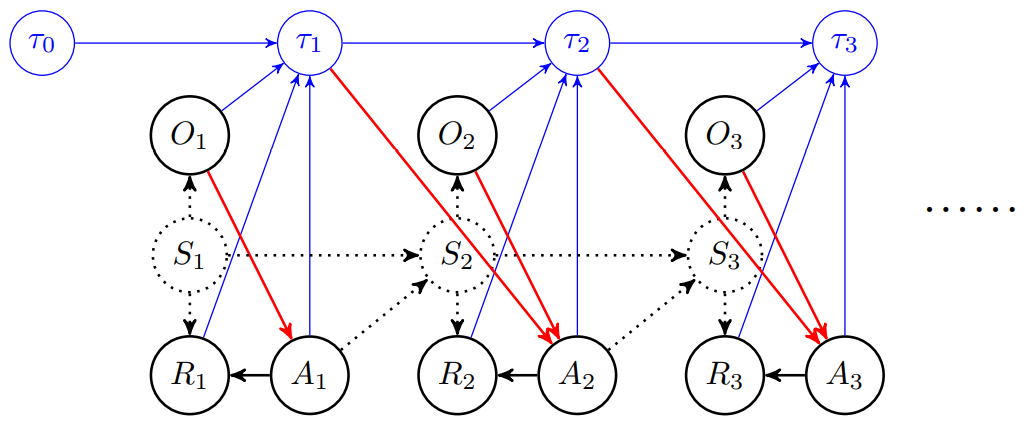}}
\caption{Examples of causal graphical illustrations of POMDP under the logging policy and the evaluation policy, where the red arrows denote the dependence relations between the policy actions $A_{t}$ and the states $S_{t}$ in (a) (while in (b) the red arrows between the policy actions $A_{t}$ and the observations $O_{t}$ and the previous trajectories $\tau_{t}$). Blue arrows represent the dependence relations of the observed trajectories~\cite{bennett2021proximal}. The randomized given logging policy with POMDP generates the true trajectories while the evaluation policy denotes the deterministic target policy that needs evaluated.}
\label{fig:bennett2021proximal}
\end{figure}
To generalize the work~\cite{tennenholtz2020off} to a non-tabular setting, Bennett et al., and Shi et al.,~\cite{bennett2021proximal,shi2022minimax} also implemented the OPE task under the POMDP model, i.e., evaluating a target policy in the POMDP given the observational data generated by some behavior policy. Bennett et al.,~\cite{bennett2021proximal} calibrated the proximal causal inference (PCI) approach to the sequential decision making setting. In particular, they formalized graphical representations under POMDP model under the logging policy and the evaluation policy.
Based on the graphs depicted in Fig.~\ref{fig:bennett2021proximal}, they then provided the identification results of evaluating the target policy via PCI, under some assumptions about the independence constraint and bridge functions. 
They finally established an estimator for evaluation. 
%
On the other hand, Shi et al.,~\cite{shi2022minimax} deduced an identification method for OPE in POMDP model in the presence of latent confounders, where the behavior policy depended on unobserved state variables while the evaluation policy only on observable observations. Their method was based on a weight bridge function and a value bridge function. Afterwards, they proposed minimax estimation methods to learn these two bridge functions, deploying function approximation techniques. 
They further proposed three estimators to learn the target policy's value, on the basis of the value function, the marginalized importance sampling and the doubly-robustness. Their proposal was suitable in settings with continuous or large observation/states spaces.

Nair et al.,~\cite{nair2021spectral} also zeroed in on the OPE problem in POMDPs, same as~\cite{tennenholtz2020off}. But unlike previous works that relied on the invertibility of certain one-step moment matrices, they relaxed such an assumption from one-step proxies to multi-step one with the past and future, motivated by spectral learning. And they developed an Importance Sampling algorithm that depended on the rank conditions on probability matrices, and not on the sufficiency conditions of observable trajectories. Their method was executed with various kinds of causal structures for proxies with the confounder.
%
The above-mentioned work including~\cite{bennett2021proximal,nair2021spectral,tennenholtz2020off} attempted to correct confounding bias in the context of POMDP model, whereas Hu et al.,~\cite{hu2022off} worked in the cases with no confounding. They proposed an OPE method to identify causal effects of a target policy also in the POMDP model, by using partial history importance weighting technique.  They further established upper and lower bounds on the error between the estimated and true values.  

As for the cooperative multi-agent settings, previous methods were faced with decentralized policies learning and multi-agent credit assignment problems. To conquer such challenges, Foerster et al.,~\cite{foerster2018counterfactual} proposed a multi-agent actor-critic algorithm called COunterfactual Multi-Agent (COMA) policy gradient, based on the POMDP model along with one more element standing for the multiple agents. In particular, their COMA majorly enjoyed three merits. Firstly, they utilized a centralized critic to estimate a counterfactual advantage for decentralized policies. Secondly, they dealt with the credit assignment problem in virtue of a counterfacutal baseline, which enabled marginalizing out a single agent's action while keeping all the other agents' actions fixed. Finally, their critic representation allowed efficient estimation for the counterfactual baseline.
%
Unlike~\cite{foerster2018counterfactual} which resorted to centralized learning, Jaques et al.,~\cite{jaques2019social} explored a more realistic scheme where each agent was trained independently. 
Specifically, they~\cite{jaques2019social} focused on reaching the goal of coordination and communication in the multi-agent RL setting. 
Their intuitive idea was to reward intrinsically to those agents who were having high causal influences on other agents' actions, where the causal influences were evaluated by counterfactual analysis. They conducted extensive experiments on two Sequential Social Dilemmas (SSDs), which were partially observed, spatially and temporally extended multi-agent games, and they showed the feasibility of their proposed social influence rewards with communication protocols. They also fostered to model other agents, allowing each agent to compute its social influence independently in a decentralized way, i.e., without access to other agents' reward functions.
%
Barton et al.,~\cite{barton2018measuring} measured the collaboration between agents via Convergent Cross Mapping (CCM), which examined the causal influence using predator agents' positions over time.

\subsubsection{Bandits}
\begin{figure}[t]
    \centering
    \includegraphics[scale=0.5]{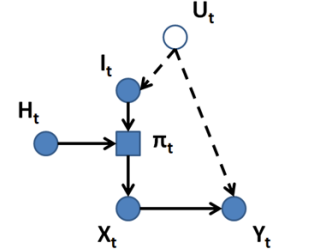}
    \caption{An example of causal graphical illustration of MAB with unobserved confounders $U_t$, where $I_t$ is the agent's intended arm choice at step $t$, $\pi_t$ denotes the policy, $X_t$ is the arm choice, $Y_t$ is the Bernoulli reward and $H_t$ denotes the agent's history~\cite{forney2017counterfactual}.}
    \label{fig:forney2017counterfactual}
\end{figure}

As for the Multi-Armed Bandit (MAB) problem, Bareinboim et al.,~\cite{bareinboim2015bandits} first showed that previous bandit methods may fall into the sub-optimal policy under the existence of unobserved confounders. Hence, they proposed a new bandit problem with unobserved confounders using causal graphical representation, where the confounder affects both the action and the outcome. 
In this setting, an observed variables $X_t \in \lbrace x_1...,x_k \rbrace$ was encoded as the agent's arm choice from $k$ arms, arms (or actions) were identified as interventions on the causal graph, and $Y_t \in \lbrace 0,1 \rbrace$ was a reward (or outcome) variable from choosing the arm $X_t$ under the unobserved confounder $U_t$. 
They hereafter developed an optimization metric employing both experimental and observational quantities and empowering Thompson Sampling~\footnote{Thompson Sampling (TS) is a traditional bandit algorithm.}, namely Causal Thompson Sampling method. Here they estimated the Effect of the Treatment on the Treated group (ETT).
%
Motivated by the unobserved confounder pitfalls in MAB from~\cite{bareinboim2015bandits}, Forney et al.,~\cite{forney2017counterfactual} employed a different vehicle for an online learning agent: counterfactual-based decision making. They first exploited counterfactual reasoning to generate counterfactual experiences by active agents. And they proposed a Thompson Sampling method where the agents combined observational, experimental and counterfactual data to learn the environments in the presence of latent confounders. It's worth noting that their confounding MAB model considered the agent's arm choice predilections and pay out rates at each round, as shown in Fig.~\ref{fig:forney2017counterfactual}.
%
Following the similar causal bandit problem as~\cite{bareinboim2015bandits}, Lattimore et al.,~\cite{lattimore2016causal} formally introduced a new causal bandit framework which subsumes traditional bandits as well as the contextual stochastic bandits. In their framework, observations were obtained after performing each intervention and the agent could control the allowed actions. To improve the learning rate at which good interventions were learned, they proposed a general algorithm to estimate the optimal action, assuming the causal graph was given (but can be arbitrary). They also proved that the algorithm enjoyed the strictly better regret than those with no exploited causal information.

To majorly improve the performance over the method~\cite{lattimore2016causal} in both theoretical and practical ways, Sen et al.,~\cite{sen2017identifying} proposed an efficient successive rejects algorithm with importance sampling, also assuming the causal structure was given a prior (so did a small part of interaction strengths). Further, they provided the first gap dependent error and simple regret bounds to identify the best soft intervention at a source variable, so as to maximize the expected reward of the target variable.
%
Since previous works only considered localized interventions~\cite{lattimore2016causal,sen2017identifying}, i.e.,  the intervention propagated locally and only to the intervened variable's neighbors, Yabe et al.,~\cite{yabe2018causal} introduced a causal bandit method via propagating inference, which achieved propagating throughout the causal graph with arbitrary interventions. In their method, the causal graph and a set of interventions were given.
They also deduced the simple regret bound, which was logarithmic with respect to the number of arms.

Harnessing a given causal graph as background knowledge, Lu et al.,~\cite{lu2020regret} proposed two algorithms, namely causal upper confidence bound (C-UCB) and causal Thompson Sampling (C-TS), to learn the optimal interventions sequentially for the bandit problem. They assumed that the distribution of parents of the reward
variable for each intervention is known. Their proposals, including the linearity cases, aimed to reduce the amount of needed explorations and enjoyed improved cumulative regret bounds. 
%
Since situations exist where performing the intervention is costly, Nair et al.,~\cite{nair2021budgeted} considered causal bandits in a given causal graph with two settings: with and without budget constraints. Such constraints determined whether to take the trade-off between costly interventions and economical observations. Hence, they proposed two algorithms, $\gamma$-NB-ALG and CRM-NB-ALG that minimized the expected simple regret and the expected cumulative regret respectively with the trade-off, utilizing causal information; and an algorithm C-UCB-2 without the budget constraint in general graphs, making the same assumption as~\cite{lu2020regret} for the distribution of reward's each parent.

Lee et al.,~\cite{lee2018where} uncovered a phenomenon that traditional strategies for MAB, including intervening on all variables at once or on all subsets of variables, would lead to sub-optimal policies. 
Hence, to guide the agent whether conducting a causal intervention or not, and which variables to intervene, they first employed the language of SCMs to characterize a new type of Multi-Armed Bandit (MAB), called SCM-MABs. They then investigated the structural properties of a SCM-MAB, which could be computed by any arbitrary causal model. Furthermore, they introduced an algorithm to identify a special set of variables, which tells the possibly-optimal minimal intervention set for the agent, allowing the existence of unobserved confounders. 
Since in many real-world applications, not all observed variables are manipulable, Lee et al.,~\cite{lee2019structural} studied such a setting of SCM-MAB problems and proposed an algorithm to identify the possibly-optimal arms. Their method was an enhanced version of the MAB algorithm, based on the crucial properties in the causal graph with manipulability constraints.
%
Lee et al.,~\cite{lee2020characterizing} zeroed in on the mixed policies which consisted of a collection of decision rules where each rule was assigned with different actions for interventions given the contexts. In particular, they first characterized the non-redundancy for the mixed polices, with a graphical criterion designed to identify the unnecessary contexts for the actions. Further they offered sufficient conditions to characterize the non-redundancy under optimality, which fostered an agent to adjust its suboptimal policy. Such characterizations lied a stark benefit to fundamentally understand the space of mixed policies and refine the agent's strategy, helping the agent fast converge to the optimum.  

As for the contextual bandit problem, Subramanian et al.,~\cite{subramanian2022causal} formalized a nuanced contextual bandit setting where the agent had the choice to perform either a targeted intervention or a standard interaction during the training phase and had access to the causal graph information. This is the first paper to introduce targeted interventions in the contextual bandit problem, and the work to integrate the causal-side information (namely causal contextual bandit problems)~\cite{subramanian2022causal}. Their proposed algorithm was based on minimizing an entropy-like measure, which reduced the demand for a large number of samples.
%
To embrace the merits of offline causal inference and online contextual bandit learning, Li et al.,~\cite{li2021unifying} proposed a framework to select some logged data to improve the accuracy of online decision making problems, including context-independent and context-dependent decisions. Given the contextual bandit model with confounders, they also derived the upper regret bounds of their forest-based algorithms. 
%
Instead of dealing with the online learning problem, Zhang et al.,~\cite{zhang2017transfer} studied the offline transfer learning problem in RL scenarios, leveraging causal inference and allowing the existence of unobserved confounders. In particular, given a causal model, they first connected algorithms to achieve identifying causal effects (from trajectories of the source agent to the target agent's action) and off-policy evaluation. In the cases where causal effects could not be identified, they proposed a new strategy to extract causal bounds, which were used to learn the policy with high learning speed. 

\begin{figure}[t]
    \centering
    \includegraphics[scale=0.15]{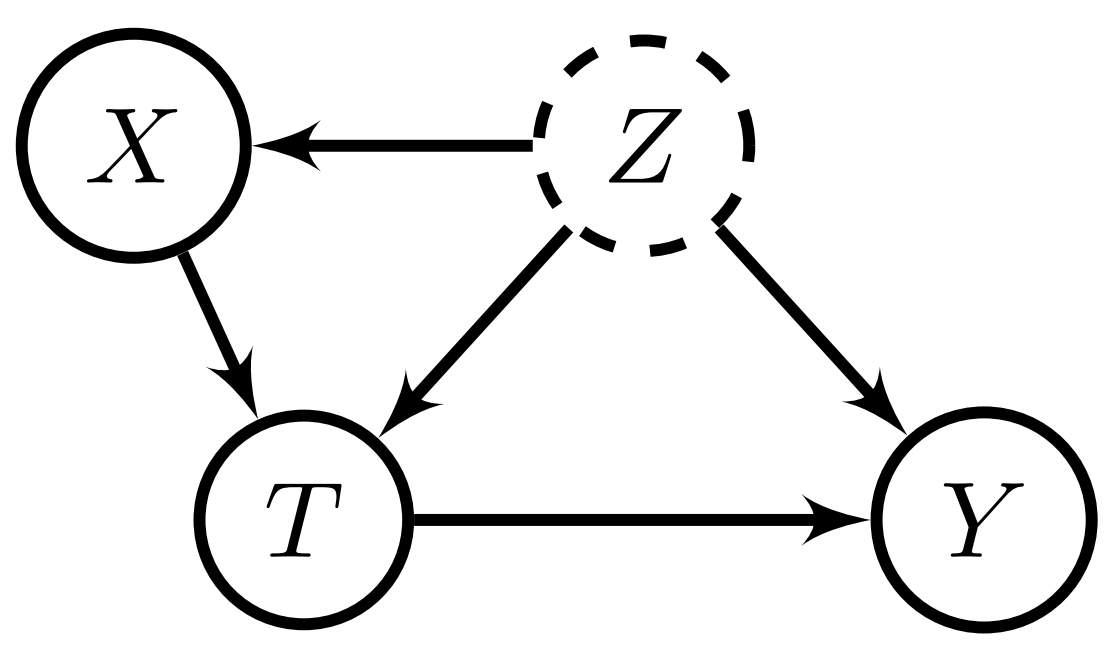}
    \caption{An example of causal graphical illustration of contextual MAB, where $X$ denotes observed covariates, $T$ is an observed treatment, $Y$ is an outcome and $Z$ stands for latent confounders~\cite{bennett2019policy}.}
    \label{fig:bennett2019policy}
\end{figure}
As for the policy evaluation problem with latent confounders, Bennett et al.,~\cite{bennett2019policy} developed an importance weighting approach based on the contextual decision making setting. Given the DAG representation of model where the confounder affected the proxies, treatment and the outcome in Fig.~\ref{fig:bennett2019policy}, they induced an adversarial objective for selecting optimal weights with some given function classes.
Such optimal weights were proved to produce consistent policy evaluations.
This OPE work with latent confounders in contextual bandit settings was considered as a special case of~\cite{bennett2021off}. 
%
Given the identical causal graph as~\cite{bennett2019policy} in Fig.~\ref{fig:bennett2019policy} where the confounder affected the proxies, treatment and the outcome, Xu et al.,~\cite{xu2021deep} proposed a Deep Feature Proxy Variable method (DFPV) to estimate the causal effect of treatment on outcomes with unobserved confounders, using proxy variables of confounding. Their proposal, which was based on the proximal causal inference (PCI), was applied to bandit OPE with continuous actions. Note that previous work~\cite{tennenholtz2020off} applied PCI as well for OPE, but with discrete action-state spaces.

Gonzalez et al.,~\cite{gonzalez2018playing} proposed a method for the agent to make decisions in an uncertain environment which was governed by causal mechanisms. 
Their method allowed the agent to hold beliefs about the causal environment, which were kept updating using interactive outcomes. In the experiments, they assumed the agent knew the causal structure of the environment. 

As for the policy improvement problem with unobserved confounders, where the confounder affected both treatments and outcomes, Kallus et al.,~\cite{kallus2018confounding} developed a robust framework to optimize the minimax regret of a candidate policy against a baseline policy, motivated by sensitivity analysis in causal inference. They optimized over the uncertainty sets with nominal propensity weights, using subgradient descent method to perform policy optimization.
Afterwards, Kallus et al.,~\cite{kallus2021minimax} provided an extended version of the above method~\cite{kallus2018confounding}, which induced sharp regret bounds in binary- as well as multiple-treatment settings. Further, they generalized theoretical guarantees on the performance of the learned policy, on the basis of previous theories~\cite{kallus2018confounding}.

\subsubsection{DTR} 
Dynamic Treatment Regime (DTR) aims at offering a set of decision rules, one per stage of intervention, that indicate what treatments are for an individual patient to improve the long-term clinical outcome, given the patient's evolving conditions and histories. 
DTR differs from the traditional MDP in that it does not require the Markov assumption and what it estimates is an optimal adaptive dynamic policy that makes its decision based on all previous information available~\cite{liao2021instrumental}.
It is an attractive vehicle for personalized management in medical research~\cite{zhang2019near}. 

Apart from Definition~\ref{def:DTR}, DTR could also be defined in detail with the language of SCM in causality~\cite{murphy2003optimal}. That is, a DTR is defined as a SCM, i.e., a tuple $<\textbf{U}, \textbf{V}, \textbf{F}, P(\textbf{u})>$, where $\textbf{U}$ represents exogenous unobserved variables (or noise variables); $\textbf{V}=\lbrace \textbf{X}, \textbf{S}, Y\rbrace$ is a set of endogenous observed variables, $\textbf{X}$ is a set of action variables, $\textbf{S}$ a set of state variables, and $Y$ is the outcome; $\textbf{F}$ represents a set of structural functions; and $P(\textbf{u})$ is the probability distribution of noises $\textbf{u}$. For stage $k=1,...,K$, a state is determined by a transition function: $s_k=\tau_k(\bar{\textbf{x}}_{k-1},\bar{\textbf{s}}_{k-1}, \textbf{u})$, a decision is decided by a behavior policy: $x_k=f_k(\bar{\textbf{s}}_{k},\bar{\textbf{x}}_{k-1}, \textbf{u})$, and an outcome ranging from 0 to 1 is determined by a reward function: $y=r(\bar{\textbf{x}}_{K},\bar{\textbf{s}}_{K}, \textbf{u})$. $\bar{\textbf{x}}_{k}$ stands for a sequence of values from variables $\lbrace X_1,...,X_k\rbrace$, i.e., $\bar{\textbf{x}}_{k} = \lbrace x_1,...,x_k\rbrace $~\cite{murphy2003optimal}.
Especially, when there is only one stage in DTR with an empty set of states, the model of such a DTR is a multi-armed bandit model.

\begin{figure}[t!]
\centering
\subfigure[]{
\includegraphics[width=0.22\textwidth]{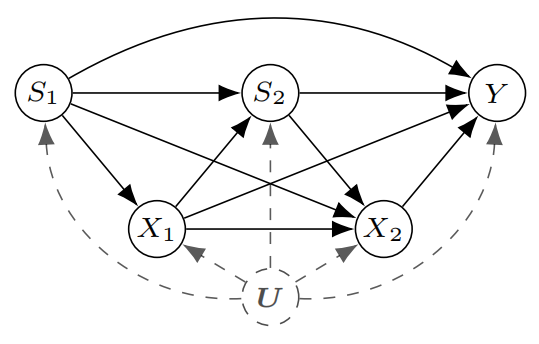}} 
\subfigure[]{
\includegraphics[width=0.22\textwidth]{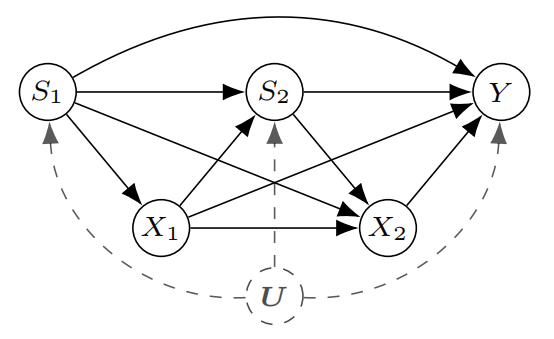}}
\caption{Examples of causal graphical illustrations of (a) a DTR with 2 stages of intervention; and (b) a DTR in (a) under sequential interventions towards $X_1$ and $X_2$: $do(X_1 \sim \pi_1(X_1|S_1), X_2 \sim P(X_2|S_1, S_2, X_1))$.  $\textbf{X}$ denotes the set of treatments, $\textbf{S}$ denotes the set of covariates, $Y$ is the outcome, and $U$ is the unmeasured confounder~\cite{zhang2019near}.}
\label{fig:zhang2019near}
\end{figure}
To learn the optimal DTR in the online RL setting, Zhang et al.,~\cite{zhang2019near} developed an adaptive RL algorithm with the near-optimal regret. Based on the DTR structure where the confounders existed as demonstrated in Fig.~\ref{fig:zhang2019near}, they derived causal bounds about the dynamics of system, which were integrated to calibrate the algorithm with accelerated and improved performance. 
Their bounds were relative to the domains of treatments $\textbf{X}$ and covariates $\textbf{S}$.

If the domains were huge, such regrets would not be attainable. Hence, to reduce the dimensionality of the policy space, Zhang et al.,~\cite{zhang2020designing} first proposed an algorithm to remove irrelevant treatments or evidence by exploiting the independence constraints in the causal structure of the DTR. Then, they designed two online RL algorithms to identify the optimal DTR in an unknown environment, by making the best of the causal order relationships in the form of a given causal diagram. One algorithm employed the principle of optimism in the face of uncertainty to deduce tight regret bounds for DTR while the other one leveraged the heuristic of posterior sampling.

Zhang et al.,~\cite{zhang2022online} took care of the mixed policy scopes, which consisted of a collection of treatment regimens resulted from all possible combinations of treatments. Specifically, they designed an online learning algorithm to identify an optimal policy with mixed policy scopes, achieving sublinear cumulative regret. Further, provided with a causal diagram, they introduced a parametrization for SCMs with finite latent states, which enabled them to parametrize all observational and interventional distributions through using the minimal number of decomposition factors, named c-components. They then proposed another practical policy learning algorithm based on the Thompson sampling, which was computationally feasible and enjoyed the merit of asymptotic similar bound on the cumulative regret. 

Chen et al.,~\cite{chen2021estimating} considered the problems of estimating and improving DTR with a time-varying instrumental variable (IV) in the case where there existed unobserved confounders affected the treatment and the outcome. They thus developed a generic identification framework, under which they derived a Bellman equation based on the classical Q-learning and defined a class of IV-optimal DTRs. They further proposed an IV-improvement framework, which enabled to strictly improve the baseline DTR under the multi-stage setting.

\subsubsection{IL}
Imitation learning (IL) is originated from the fact that children learn skills via mimicking adults, which therefore aims at learning imitating policies from demonstrations generated by an expert~\cite{hussein2017imitation,osa2018algorithmic}. Existing IL methods roughly fall into two categories, namely behavior cloning (i.e., directly mimicking the behavior policy of an expert) and inverse reinforcement learning (i.e., learn a reward function from expert's trajectories and apply RL method for the policy). Here, we start to survey some IL methods along with causality.

To endow the agent to obtain imitability to learn an optimal policy, Zhang et al.,~\cite{zhang2020causal} started with SCMs and utilized causal languages to propose a method to learn an imitating policy from combinations of demonstration data, allowing the existence of unobserved confounders. Their method took a causal diagram, a policy space and an observational distribution as inputs, with a complete graphical criterion for determining the imitation feasibility. The demonstrator and the learner might have different observations.
%
\begin{figure}[t!]
\centering
\subfigure[Source domain]{
\includegraphics[width=0.22\textwidth]{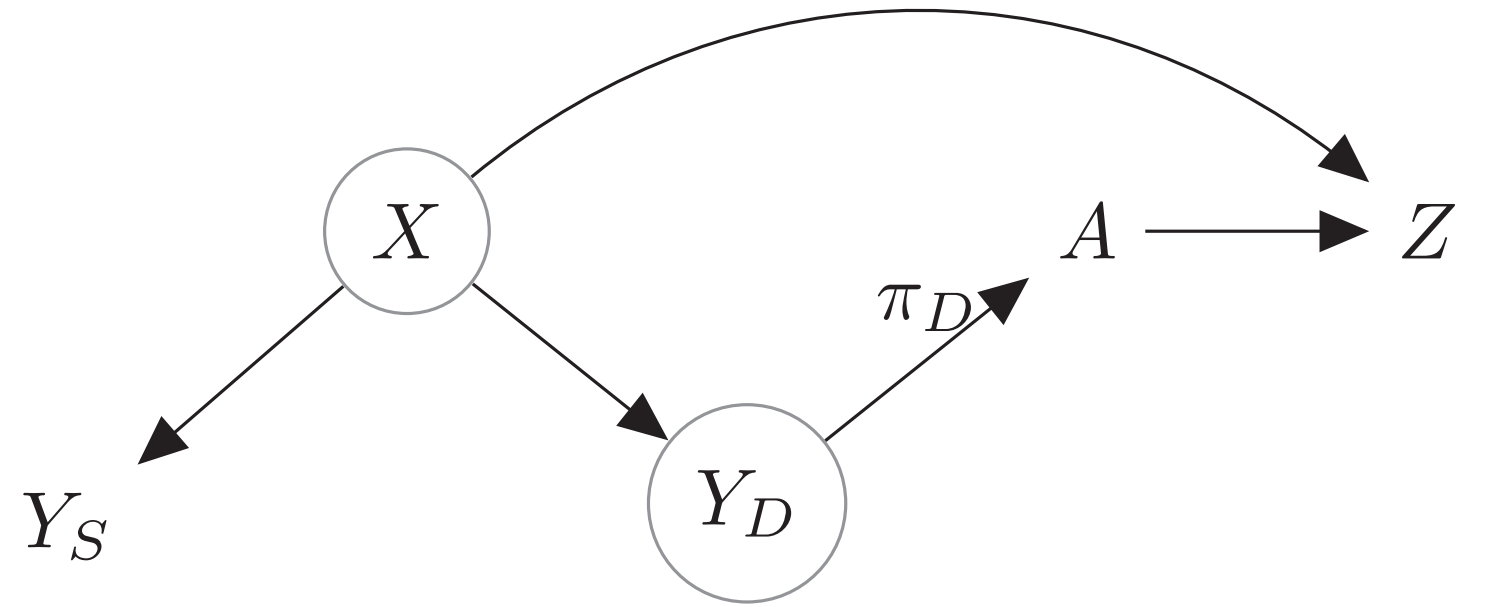}} 
\subfigure[Target domain]{
\includegraphics[width=0.19\textwidth]{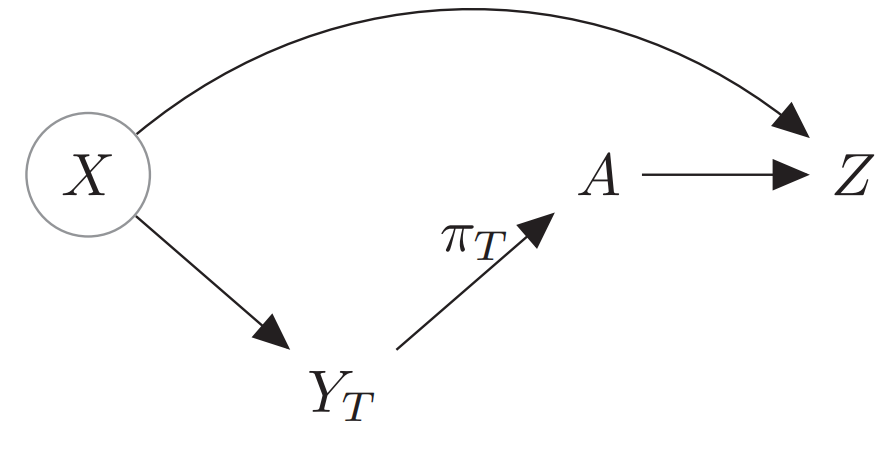}}
\caption{Examples of causal graphical illustrations of IL in the (a) source domain; and (b) target domain.  
${X}$ denotes the state, $A$ denotes the action, and $Z$ is the outcome. $Y_D$ stands for the demonstrator’s input, generated by the demonstrator’s sensors, while $Y_S$ the spectator’s observation of the state of the source system. $Y_T$ denotes in the target domain the input from the target agent’s sensors~\cite{etesami2020causal}.}
\label{fig:etesami2020causal}
\end{figure}
Similarly, Etesami et al.,~\cite{etesami2020causal} focused on addressing the differences and shifts between (a) the demonstrator's sensors, (b) our sensors and (c) the agent's sensors. According to the known causal graphs between states, actions, outcomes and observations in source and target domains as depicted in Fig.~\ref{fig:etesami2020causal}, they developed algorithms to identify effects of the demonstrator's actions and do imitation learning. They also provided theoretical guarantees for the estimated policy. 
%
Tennenholtz et al.,~\cite{tennenholtz2022covariate} handled the problem of using expert data with unobserved confounders and possible covariate shifts for imitation learning and reinforcement learning. They first defined a contextual MDP setup, which implied its underlying causal structure. In the IL setting (without access to rewards), they claimed that in the case where the reward depends on the context, ignoring the covariate shifts would induce catastrophic errors and render non-identifiability of model. In the RL setting, they proved that when the reward is provided, the optimality could be reached with the expert data and even arbitrary hidden covariate shifts. Eventually they proposed an algorithm based on the corrective trajectory sampling with convergence guarantees. 

The above solutions~\cite{zhang2020causal,etesami2020causal,tennenholtz2022covariate} however were limited by the single-stage decision-making setting. To explore the mismatch case in sequential settings where the learner must take multiple actions per episode, Kumor et al.,~\cite{kumor2021sequential} first induced a sufficient and necessary graphical criterion for determining the imitation feasibility, based on a known causal structure with domain knowledge of the covariates and targets. Thereafter, they proposed an efficient algorithm to optimize the policy for each action, while determining the imitability. They also proved the completeness of the criterion and verified the favorable performance of their method. 

\begin{figure}[t]
    \centering
    \includegraphics[scale=0.35]{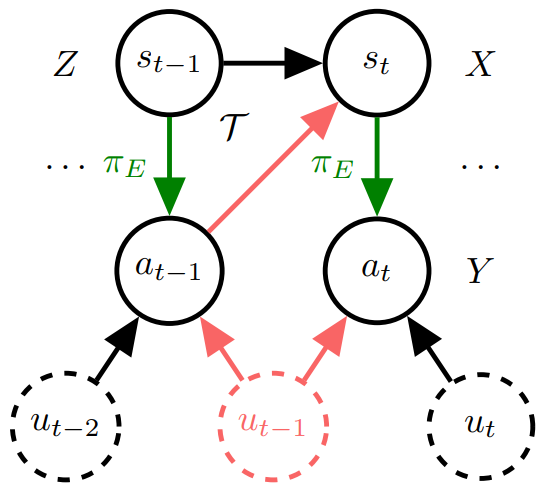}
    \caption{An example of causal graphical illustrations of IL that captures the temporally correlated noises, where such noises $u_t$ are the latent confounding variables that induce spurious correlations between states $s_t$ and actions $a_t$~\cite{swamy2022causal}.}
    \label{fig:swamy2022causal}
\end{figure}

Empirically, we might encounter the data which were corrupted by temporally correlated noises, producing spurious correlations between states and actions and resulting in unsatisfactory policy performance, as illustrated in Fig.~\ref{fig:swamy2022causal}. To tackle this issue, Swamy et al.,~\cite{swamy2022causal} leveraged variants of instrumental variable regression methods, taking the past states as instruments, and presented two algorithms to eliminate the spurious correlations. Dealing with the confounding problem based on the SCM, in IL, their methods derived consistent policies. 

Compared with traditional IRL methods, those based on the maximum causal entropy enjoyed some merits, such as less specific assumptions on the expert behavior or good performances with a wide variety of reward functions, etc~\cite{ziebart2010modeling,ziebart2013principle}. Such a causal entropy allows to describe some structured qualities for the policy.
However, these methods have to restrict to the finite horizon setting. To this end, Zhou et al.,~\cite{zhou2017infinite} extended the maximum causal entropy framework to infinite time horizon setting. They considered the maximum discounted causal entropy as well as the maximum average causal entropy, and proposed a gradient-based algorithm with the stationary soft Bellman policy.

\subsection{CRL with Unknown Causal Information} \label{sec:unknown}
In this subsection, we take a review on CRL methods where the causal information is unknown and has to be learned in advance. 
Compared with the first category, it is more challenging. And methods usually contain two phases: causal information learning and policy learning, which are carried out iteratively or in turn.
Regarding the policy learning phase with causality, the paradigm is consistent with that of the first category.
Regarding the causal information learning, it usually involves the causal structure learning (that may include latent confounders, or causes of actions or rewards), or causal representation/feature/abstraction learning. Most of these CRL methods adopt casual discovery techniques (including variants of PC~\cite{Pearl2000causality}, CD-NOD~\cite{huang2020causal}, Granger~\cite{granger1969investigating}, etc),  intervention, deep neural networks (including Variational Auto-Encoder~\cite{kingma2013auto}, etc) for causal information learning. 

We summarize these methods according to different model settings, i.e., MDP, POMDP, bandits and IL.
\begin{table}[ht]	\label{tab:summary_unknown}
	\caption{Part of CRL algorithms with unknown causal information}
	\begin{center} 
	\begin{tabular}{  m{1cm} m{7cm} } 
	    \toprule
		\multicolumn{1}{c}{\bf Models}  &\multicolumn{1}{c}{\bf Algorithms} \\ 
		\midrule
		MDP & ICIN~\cite{nair2019causal}, FOCUS~\cite{zhu2022offline}, causal InfoGAN~\cite{kurutach2018learning}, NIE~\cite{Herlau2022reinforcement}, DBC~\cite{zhang2021learning}, IBIT~\cite{mozifian2020intervention}, CoDA~\cite{pitis2020counterfactual}, CAI~\cite{seitzer2021causal}, FANS-RL~\cite{feng2022factored}, CDHRL~\cite{peng2022causalitydriven}, CDL~\cite{wang2022causal}, etc.\\
		\midrule
		POMDP & causal curiosity~\cite{sontakke2021causal}, Deconfounded AC~\cite{lu2018deconfounding}, RL with ASR~\cite{huang2021action}, AdaRL~\cite{huang2022adarl}, causal states~\cite{zhang2019learning}, IPO~\cite{sonar2021invariant}, etc.\\
		\midrule
		Bandits  & CN-UCB~\cite{lu2021causal}, Multi-environment Contextual Bandits~\cite{saengkyongam2021invariant}, Linear Contextual Bandits~\cite{tennenholtz2021bandits}, etc.  \\
		\midrule
		IL  & CIM~\cite{samsami2021causal}, CPGs~\cite{katz2017cognitive}, causal misidentification~\cite{de2019causal},  CIRL~\cite{bica2020learning}, CILRS~\cite{codevilla2019exploring}, ICIL~\cite{bica2021invariant}, cause-effect IL~\cite{katz2017novel}, copycat~\cite{wen2020fighting,chuang2022resolving,wen2021keyframe,wen2022fighting}, etc.
		\\
		\bottomrule
	\end{tabular}
	\end{center}
\end{table}

\subsubsection{MDP}
To show that a causal world-model can surpass a plain world-model for the offline RL in both theoretical and practical ways, Zhu et al.,~\cite{zhu2022offline} first proved generalization error bounds of the model prediction and policy evaluation with causal and plain world-models, and proposed an offline algorithm called FOCUS based on the MDP. FOCUS aimed at learning the causal structure from data by the extended PC causal discovery methods, and leveraging such a learned structure to optimize a policy via an offline MBRL algorithm, MOPO~\cite{yu2020mopo}.

Méndez-Molina~\cite{ijcai2021-684} presented in his research proposal that there existed certain characteristics in robotics that accelerated causal discovery via interventions. Hence, he aimed at providing an intelligent agent that could simultaneously identify causal relationships as well as learn policies. Based on MDPs, he took inspiration of the Dyna-Q algorithm, and after discovering the underlying causal structure as the model, he used it to learn a policy for a given task by a model-free method.
%
To capture causal knowledge in the RL setting, Herlau et al., ~\cite{Herlau2022reinforcement} introduced a causal variable in the DOORKEY environment, which was served as a manipulatable mediating variable to predict the reward from the policy choice. Such a causal variable was binary, and was identified by maximizing the natural indirect effect (NIE) in mediation analysis. Thereafter, they built a small causal graph between this causal variable, policy choice and the return so as to help optimize the policy.
%
To enable the agents to complete goal-conditioned tasks with causal reasoning capability, Nair et al.,~\cite{nair2019causal} formalized a two-phase meta-learning-based algorithm. In the first phase, they trained causal induction models from visual observations, which induced a causal structure via an agent's interventions; in the second phase, they leveraged the induced causal structure for a goal-conditioned policy with an attention-based graph enconding. Their method improved generalization abilities towards unseen causal relations as well as new tasks in new environments.  
%
Aiming at goal-directed visual plans, Kurutach et al.,~\cite{kurutach2018learning} combined the generalization merit from deep learning of dynamics models with the effective representation reasoning from classical planners, and developed a causal InfoGAN framework to learn a generative model of high-dimensional sequential observations. Via such a framework, they obtained a low-dimensional representation which stood for the causal nature of data. With this structured representation, they employed planning algorithms for goal-directed trajectories, which were then transformed into a sequence of feasible observations.
%
Ding et al.,~\cite{ding2022generalizing} augmented Goal-Conditioned RL with Causal Graph (CG), that characterized causal relations between objects and events. They first used interventional data to estimate the posterior of CG; and then used CG to learn generalizable models and interpretable policies.

For complex tasks with sparse rewards and large state spaces, it is hard for Hierarchical Reinforcement Learning (HRL) to discover high-level hierarchical structures to guide the policy learning, due to the low exploration efficiency. To leverage the advantages from causality, Peng et al.,~\cite{peng2022causalitydriven} proposed a Causality-Driven Hierarchical Reinforcement Learning (CDHRL) framework to achieve effective subgoal hierarchy structures learning. 
CDHRL contains two processes: causal discovery and hierarchy structures learning, which are boosted by each other. 
Building discrete environment variables in advance in the subgoal-based MDP model, they adopted an iterative way to progressively discover the causal structure between such environment variables, and construct the hierarchical structure where reachable subgoals are those controllable environment variables from the discovered causality. 
They evaluated their framework's efficiency in both causal discovery and hierarchy construction in complex tasks, i.e., 2d-Minecraft~\cite{sohn2018hierarchical} and simplified
sandbox survival games Eden~\cite{chen2021eden}.
%

It is hard for traditional methods to learn the minimal causal models of many environments. Furthermore, causal relationships in most environments may vary at every time step, termed as the time-variance problem~\cite{luczkow2021structural}. To address such a problem, Luczkow~\cite{luczkow2021structural} introduced a local representation $(f,g)$ in an MDP that satisfies
\begin{equation}
    s_{t+1} = f(s_t, a_t, g(s_t, a_t)),
\end{equation}
where $f$ is a function representing the causal mechanism from the causes to the next state $s_{t+1}$, $a_t$ is the action at time $t$, and $g(\cdot)$ is a function predicting the causal structure between states at each time step. Following this model formalization, they accomplished learning both causal models at the time step and the transition function, which could be used to planning algorithms and produced a policy.
%
To extend existing stationary frameworks~\cite{huang2022adarl} to the non-stationarity settings where transitions and rewards could be time-varying within an episode or cross episodes in environments, Feng et al.,~\cite{feng2022factored} proposed a Factored Adaptation algorithm for non-stationary RL (FANS-RL), based on the factored non-stationary MDP model. FANS-RL aimed at learning a factored representation that encoded temporal changes of dynamics and rewards, including continuous and discrete changes. The learned representation then could be integrated with SAC~\cite{haarnoja2018soft} for identifying the optimal policy.
%

\begin{figure}[t]
    \centering
    \includegraphics[scale=0.45]{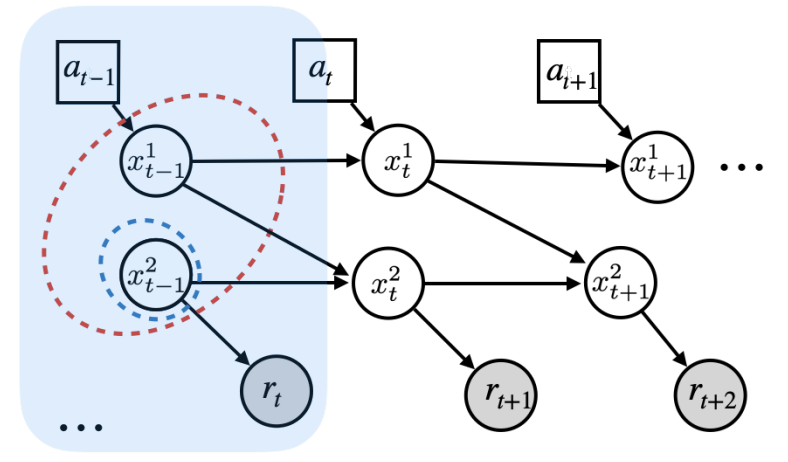}
    \caption{An example of causal graphical illustration of MDP, with states $s=\lbrace x^1, x^2 \rbrace$ when ignoring timesteps~\cite{zhang2020invariant}. Note that $x^2$ is the only parent of the reward $r$.}
    \label{fig:zhang2020invariant}
\end{figure}
To improve generalization across environments of RL
algorithms, Zhang et al.,~\cite{zhang2020invariant} focused on learning the state abstraction or representation from the varying observations. Based on block MDPs shown in Fig.~\ref{fig:zhang2020invariant}, where sets of environments with a shared latent state space and dynamics structure over that latent space, they proposed a method using invariant causal prediction to learn the model-irrelevance state abstractions, which enable to generalize across environments with a shared causal structure. The model-irrelevance state abstraction in Fig.~\ref{fig:zhang2020invariant} must include both variables $x^1, x^2$.
%
Based on the sparsity property in the dynamics that each next state variable only depends on a small subset of current state and action variables, Tomar et al.,~\cite{tomar2021modelinvariant} proposed a model-invariant state abstraction learning method. Within the MBRL setting, their method allowed for the generalization to unseen states.
To further learn invariant representations for effective downstream control, Zhang et al.,~\cite{zhang2021learning} proposed a method called Deep Bisimulation for Control (DBC) to learn robust latent representations which encoded task-relevant information, using bisimulation metrics without reconstruction. Intuitively, their focus was on the components in the state space that causally affected current and future rewards. They also proved the value bounds between the optimal value function of the true MDP and the one of the constructed MDP with the learned representations. 
%
Though Zhang et al.,~\cite{zhang2021learning} learned the latent representations encoding the task-relevant information, they might elide the information which was irrelevant to the current task but relevant to another task, and they assumed the dynamics model between state abstractions was dense. To handle these issues, Wang et al.,~\cite{wang2022causal} proposed a method of learning causal dynamics for task-independent state abstraction. Specifically, they first decomposed state variables into three types: controllable variables, action-relevant variables and action-irrelevant variables. They then conducted conditional independence tests based on the conditional mutual information to identify the causal relationships between variables. By removing those unnecessary dependencies, they learned the dynamics that generalized well to unseen states, and finally derived the state abstraction.
%
To generalize to unseen domains for the RL agents, Mozifian et al.,~\cite{mozifian2020intervention} presented Intervention-Based Invariant Transfer learning algorithm (IBIT) in a multi-task setting, which regarded the domain randomization and data augmentation as forms of interventions to take away the irrelevant features from pixel observations. They derived an invariance objective via bisimulation matrics to learn the minimal causal representation of the environment which was relative to the relevant features of the solving task. Their method improved the zero-shot transfer performances of a robot arm in sim2sim as well as sim2real settings.
%
To facilitate the agent to distinguish between causal relationships and spurious correlations in the environment of online RL, Lyle et al.,~\cite{lyle2021resolving} proposed a robust target-then-execute exploration algorithm to identify causal structures in MDP models, with theoretic motivations about the unbiased estimates of the value of any state-action pair.    

Following the idea of meta-learning, Dasgupta et al.,~\cite{dasgupta2019causal} used the model-free RL algorithm (A3C) to derive a method which enabled causal reasoning. They trained the agents to implement causal reasoning in three settings: observational, interventional and counterfactual, to gain rewards. This was the first algorithm that took a fusion between causal reasoning and meta model-free reinforcement learning.

\begin{figure}[t]
    \centering
    \includegraphics[scale=0.35]{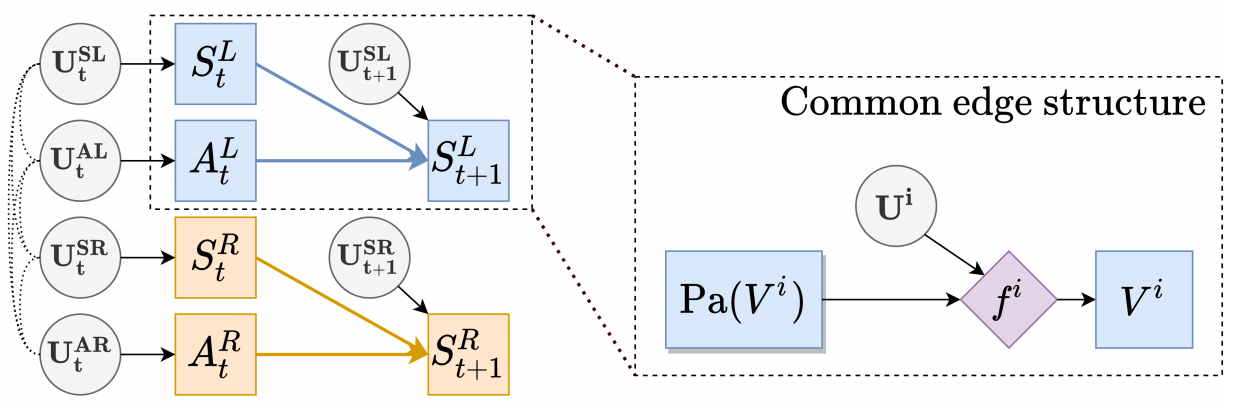}
    \caption{An example of causal graphical illustration of MDP, where these state and action spaces were decomposed into two locally independent subspaces: $\mathcal{S} = \mathcal{S}^L\oplus \mathcal{S}^R, and \mathcal{A} = \mathcal{A}^L \oplus \mathcal{A}^R$~\cite{pitis2020counterfactual}.}
    \label{fig:pitis2020counterfactual}
\end{figure}
In many scenarios like the control of a two-armed robot or a game of billiards, there may exist some sparse interactions where dynamics can be decomposed to locally independent causal mechanisms, as demonstrated in Fig.~\ref{fig:pitis2020counterfactual}. Based on this finding, Pitis et al.,~\cite{pitis2020counterfactual} first introduced local causal models, which could be obtained by conditioning a subset of the state space from the global model. They then proposed an attention-based approach to discover such models for Counterfactual Data Augmentation (CoDA). Their method finally improved the sample efficiency for policy learning in batch-constrained and goal-conditioned settings. 
%
Different from~\cite{pitis2020counterfactual} which used local causal models to augment the data counterfactually for the policy training, Seitzer et al.,~\cite{seitzer2021causal} used local causal models to detect situation-dependent causal influence, and to improve exploration and off-policy learning. In accordance with the intuition that only when the robot contacted the object, could the object be moved, they derived a measure of such local Causal Action Influences (CAI) with conditional mutual information. Integrating this measure into RL algorithms, their proposed CAI acted as an exploration bonus to perform active action exploration and to prioritize in experience replay for off-policy training.

In MBRL, there exist some endeavors that zero in on building the partial models with only part of the observatiosns. Considering the partial models, Rezende et al.,~\cite{rezende2020causally} pointed out that previous works may not be causally correct, since part of the modeled observations may be confounded by other unmodeled observations. Hence, with MDP, they tried to figure out the information of states which affected the policy but belonged to unmodeled observations, and viewed such information as confounders. They proposed to learn a family of modified partial models for RL. 

Transfer learning lays the groundwork for RL to deal with the data inefficiency and model-shift issues, but it may still suffer from the interpretability problem. Sun et al.,~\cite{sun2021model} approached such a problem and presented a model-based RL algorithm which achieved explainable and transferable learning via causal graphical representations. The causal model aided in encoding the invariant and changing modules across domains. Specifically, based on the augmented source domain data, they firstly used a feasible causal discovery method, CD-NOD~\cite{huang2020causal}, to discover the causal relationships between variables of interest. Then, they leveraged the learned augmented DAG to infer the target model by training an auxiliary-variable neural network. They eventually transferred the learned model to train a target policy employing DQN or DDPG. 
%
To handle zero-shot multi-task transfer learning, Kansky et al.,~\cite{kansky2017schema} introduced Schema Networks that are generative causal models for object-oriented reinforcement learning and planning. These networks allowed to explore multiple causes of events and
reason backward through causes to achieve goals, with training efficiency and robust generalization.


\subsubsection{POMDP}
To bridge the gap between RL and causality, Gasse et al.,~\cite{gasse2021causal} cast the model-based RL as a causal inference model. Specifically, they made the best of both online interventional and offline observational data on the basis of POMDP, and they proposed a generic method, aimed at uncovering a latent causal transition model, so as to infer the POMDP transition model via deconfounding. They also gave the correctness and efficiency proofs in the asymptotic case. 
%
%
Lu et al.,~\cite{lu2018deconfounding} considered the RL problems which utilized the observational historical data to estimate policies when confounders might exist. They named the formalization as "deconfounded reinforcement learning" and also extended existing actor-critic algorithms to their deconfounding variants. Roughly speaking, they first identified a causal model from the purely observational data, discovering the latent confounders as well as estimating their causal effects on the action and reward. They thereafter deconfounded the confounders with the causal knowledge. Based on the learned and deconfounded model, they optimized the final policy. This was the first endeavour to combine confounders cases and the full RL algorithms.

\begin{figure}[t]
    \centering
    \includegraphics[scale=0.45]{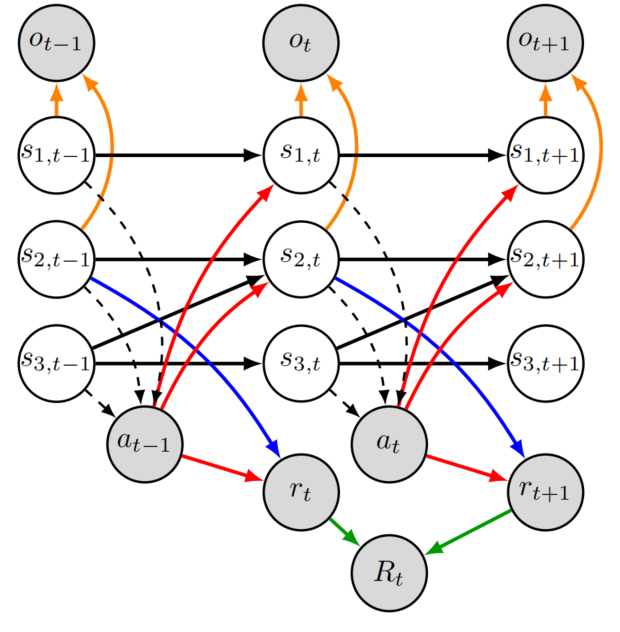}
    \caption{An example of causal graphical illustration of POMDP, where grey nodes are observed variables while white nodes are unobserved ones~\cite{huang2021action}.}
    \label{fig:huang2021action}
\end{figure}

Considering that perceived high-dimensional signals might contain some irrelevant or noisy information for decision-making problems in real-world scenarios, Huang et al.,~\cite{huang2021action} introduced a minimal sufficient set of state representations (termed Action-Sufficient Representations, ASRs for short) that captured sufficient and necessary information for downstream policy learning. In particular, they first established a generative environment model which characterized the structural relationships between variables in the RL system, as illustrated in Fig.~\ref{fig:huang2021action}. And by constraining such relationships and maximizing the cumulative reward, they presented a structured sequential Variational Auto-Encoder method to estimate the environment model as well as the ASRs. This learned model and ASRs then would accelerate the policy learning procedure effectively. 

Following the similar construction of the environment model in Fig.~\ref{fig:huang2021action}, Huang et al.,~\cite{huang2022adarl} proposed a unified framework called AdaRL for the RL system to make reliable, efficient and interpretable adaptions to new environments. Specifically, they made the best of their proposed graphical representations to characterize what and where the changes adapt from the source domains to the target domain. 
That is, via learning factored representations that encoded changes in observation, dynamics and reward functions, they were capable of learning an optimal policy from the source domains. Thereafter, such a policy was adapted in the target domain with the identified changes. 
Experimental results demonstrated their superior policy adaptation performances with graphical representations.
%
Sonar et al.,~\cite{sonar2021invariant} deployed an invariance principle to propose a reinforcement learning algorithm termed as the invariant policy optimization, with generalization ability beyond training domains. Following the POMDP setting in each domain, they learned a representation $\Phi$ mapping observations $o_t \in \mathcal{O}$ to hidden variables $h_t \in \mathcal{H}$ such that there existed a policy $\pi$ mapping $h_t$ to $a_t \in \mathcal{A}$ which was simultaneously optimal across domains. They finally obtained the invariant policy during training, which enabled generalization since such a policy found the causes of successful actions intuitively.
%

Aiming at the problems of data-efficiency and interpretability, Liang et al.,~\cite{liang2020learning,liang2021inferring} presented an algorithm for POMDPs to discover time-delayed causal relations between events from a robot at regular or arbitrary times, and to learn transition models. 
They introduced hidden variables to stand for past events in memory units and to reduce stochasticity for observations. In such a way, they achieved training a neural network to predict rewards and observations. while identifying a graphical model between observations at different time steps.
They then used planning methods for the model-based RL.
%
Motivated by the fact that animals could reason about their behaviors through interactions with the environment, and discern the causal mechanism for the variations, Sontakke et al.,~\cite{sontakke2021causal} first proposed a causal POMDP model and introduced an intrinsic reward, called causal curiosity. This causal curiosity enabled the agent to identify the causal factors in the dynamics of the environment, giving interpretation about the learned behaviors and improving sample efficiency in transfer learning.  
%
To enable efficient policy learning in the environments with high-dimensional observations spaces, Zhang et al.,~\cite{zhang2019learning} proposed a principled approach to estimate the causal states of the environment, which coarsely represented the history of actions and observations in POMDPs. Such causal states formed a discrete causal graph and helped predict the next observation given the history, which further facilitated the policy learning.


\subsubsection{Bandits}
As for the Multi-Armed Bandit (MAB) problem in the causal discovery setting, Lu et al.,~\cite{lu2021causal} proposed a causal bandit algorithm, namely Central Node UCB (CN-UCB), without knowing the causal structure. Their method was suitable for causal trees, causal forests, proper interval graphs, etc. Particularly, they first took an undirected tree skeleton structure as input, and output the direct cause of the reward. Then they applied the UCB algorithm on the reduced action set to identify the best arm. They finally showed theoretically that under some mild conditions, their method's regret scaled logarithmically with the number of variables in the causal graph. As stated, their method was the first causal bandit approach with better regret guarantee than standard MAB methods, in cases where the causal graph was not known in prior. 

As for the contextual bandit problem, Saengkyongam et al.,~\cite{saengkyongam2021invariant} solved the environmental shift issue in the offline setting, via the lens of causality. Their proposed multi-environment contextual bandit algorithms allowed for changes in the underlying mechanisms. They also delineated that in the case where unobserved confounders were present, the policy invariance they introduced could foster the generalization across environments under some assumptions. The heart of the invariance policy lied at the identification of the invariance set, which could be implied by the structure learning or be tested by the statistical method under distributional shifts~\cite{thams2021statistical}.
%
Tennenholtz et al.,~\cite{tennenholtz2021bandits} studied the problem of linear contextual bandits with observed offline data. They showed that the partially observed confounded information, which was characterized as linear constraints, could be estimated and utilized to online learning. Their proposed method could achieve the better overall regret.

\subsubsection{IL}

\begin{figure}[t]
    \centering
    \includegraphics[scale=0.3]{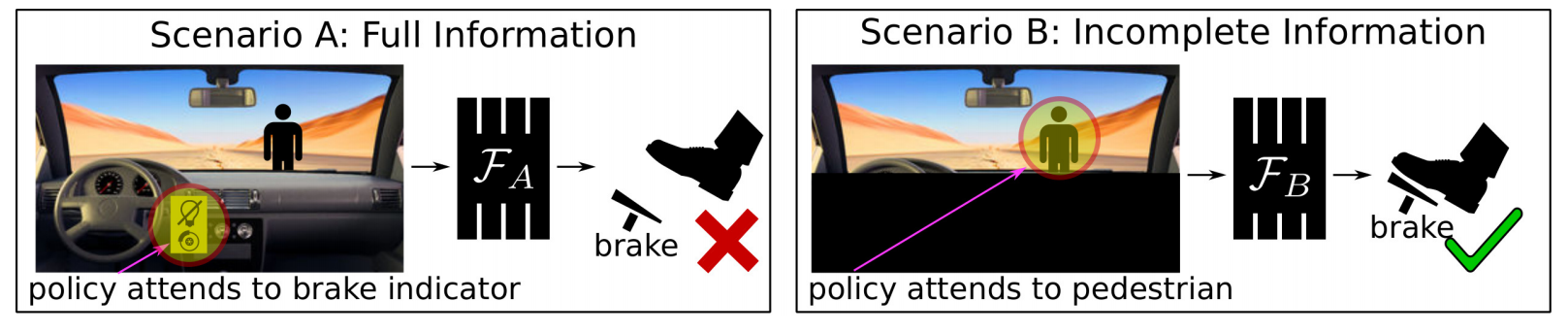}
    \caption{Causal misidentification phenomenon: access to more information may yield worse imitation learning performance~\cite{de2019causal}.}
    \label{fig:de2019causal}
\end{figure}
Taking the car driving as an example, De et al.,~\cite{de2019causal} demonstrated that being oblivious to causality in the training procedure of IL would be damaging in learning policies, leading to a problematic phenomenon called "causal misidentification", as depicted in Fig.~\ref{fig:de2019causal}. Such a phenomenon implied that: access to more information can yield worse performance. To solve it, they proposed a causally motivated method. Firstly, they learned a mapping function from causal graphs to policies. Then through targeted interventions, either environment interaction or expert queries, they determined the correct causal models as well as policies. 
%
However, this algorithm~\cite{de2019causal} had high complexity especially in autonomous driving. Samsami et al.,~\cite{samsami2021causal} hence proposed an efficient Causal Imitative Model (CIM) to handle inertia and collision problems from autonomous driving. Such undesirable problems were resulted from ignoring the causal structure of expert demonstrations. CIM first identified causes from the latent learned representation, and using Granger causality estimated the nest position. On the basis of the causes, CIM learned the final policy for the driving car.
%
Wen et al.,~\cite{wen2020fighting,wen2021keyframe,wen2022fighting,chuang2022resolving} considered a more specific causal confusion phenomenon, "copycat problem": the imitator tends to simply copy and repeat the previous expert action at the next time step.
To combat this problem, they~\cite{wen2020fighting} proposed an adversarial IL approach to learn a feature representation that ignores the nuisance correlate from the previous action, still remaining the useful information for the next action; Wen et al.,~\cite{wen2021keyframe} gave up-weights to those keyframes which correspond to the expert action changepoints in the IL objective function, to learn copycat shortcut policies; Wen et al., proposed PrimeNet algorithm to improve rbustness and avoid shortcut solution.
%
Since it was still an unresolved problem to well scale existing autonomous driving methods to real-world scenarios, Codevilla et al.,~\cite{codevilla2019exploring} explored the limitations of behavior cloning and provided a new benchmark for investigations. For instance, they reported the generalization issue was partly due to the lack of a causal model. Data bias may lead to the inertia problem, suffering from causal confusion. Explicitly learning and employing a causal model is desirable in autonomous driving.
%
%
Volodin et al.,~\cite{volodin2020resolving} and Lee et al.,~\cite{lee2021causal} learned causal structures between the states, and actions via interventions in the environment, enabling the policy learning procedure to take only cause states as inputs and preventing the spurious correlations of variables.
\begin{figure}[t!]
    \centering
    \includegraphics[scale=0.4]{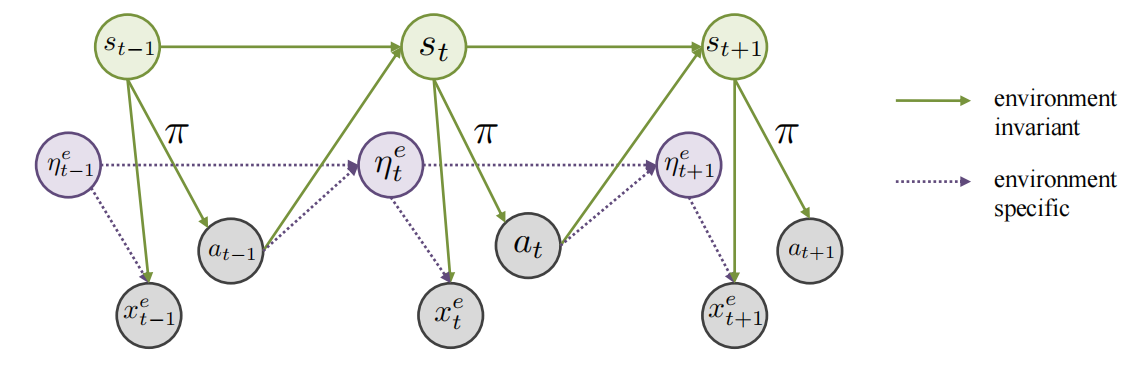}
    \caption{An example of casual graphical illustration of IL, where expert demonstrations include observations $x_t$ and actions $a_t$. States $s_t$ consist of causal parents of the actions while $\eta_t$ are noise representations that encapsulate any spurious correlations with the actions~\cite{bica2021invariant}.}
    \label{fig:bica2021invariant}
\end{figure}

As for the generalization problems, Lu et al.,~\cite{lu2022invariant} took a unified way to give the formalization, calling them the representation generalization, policy generalization and dynamics generalization. Following the behavior cloning approach to learn an imitation policy, they proposed a general framework to handle these generalization problems, with milder assumptions and theoretical guarantees. They identified the direct causes of a target variable and used such causes to predict an invariant representation which enabled generalizations.
%
Bica et al.,~\cite{bica2021invariant} aimed at learning generalizable imitation policies in the strictly batch setting from multiple environments, deploying to an unseen environment. To tackle the problems of spurious correlations between variables from expert's demonstrations, and of transition dynamics mismatch from multiple different environments, they learned a shared invariant representation, i.e., a latent structure that encoded the causes of the expert action, as well as a noise representation $\eta$ which was environmentally specific, as demonstrated in Fig.~\ref{fig:bica2021invariant}.
%

As for the explanation problems, Bica et al.,~\cite{bica2020learning} bridged the gap between counterfactual reasoning and offline batch inverse RL, and proposed to learn the reward function with regard to the tradeoffs of the expert's actions. Their approach parameterized the reward functions according to the \textit{what if} scheme, producing interpretable descriptions of sequential decision-making. Since they estimated the casual effects of different actions, counterfactuals could handle the off-policy nature of policy evaluation in the batch setting.
%
As for IL methods with unknown causal structures, a general-purpose cognitive robotic IL framework centered on the cause-effect reasoning was first proposed by Katz et al.~\cite{katz2016imitation} in 2016. This framework focused on inferring a hierarchical representation of a demonstrator's intentions with causal reasoning methods, offering an explanation about the demonstrator's actions. In such a causal interpretation, top-level intentions, represented as the abstraction of skills, then could be reused to fulfill a specific plan in new situations~\cite{katz2016imitation,katz2017novel}.
Further, during the planning procedure, Katz et al.,~\cite{katz2017cognitive} established causal plan graphs (CPGs) which modeled the causal relationships between hidden intentions, actions and goals. CPGs automatically gave causally-driven explanations of planned actions.

\subsection{Summary}
\begin{figure}[ht!]
    \centering
    \includegraphics[scale=0.4]{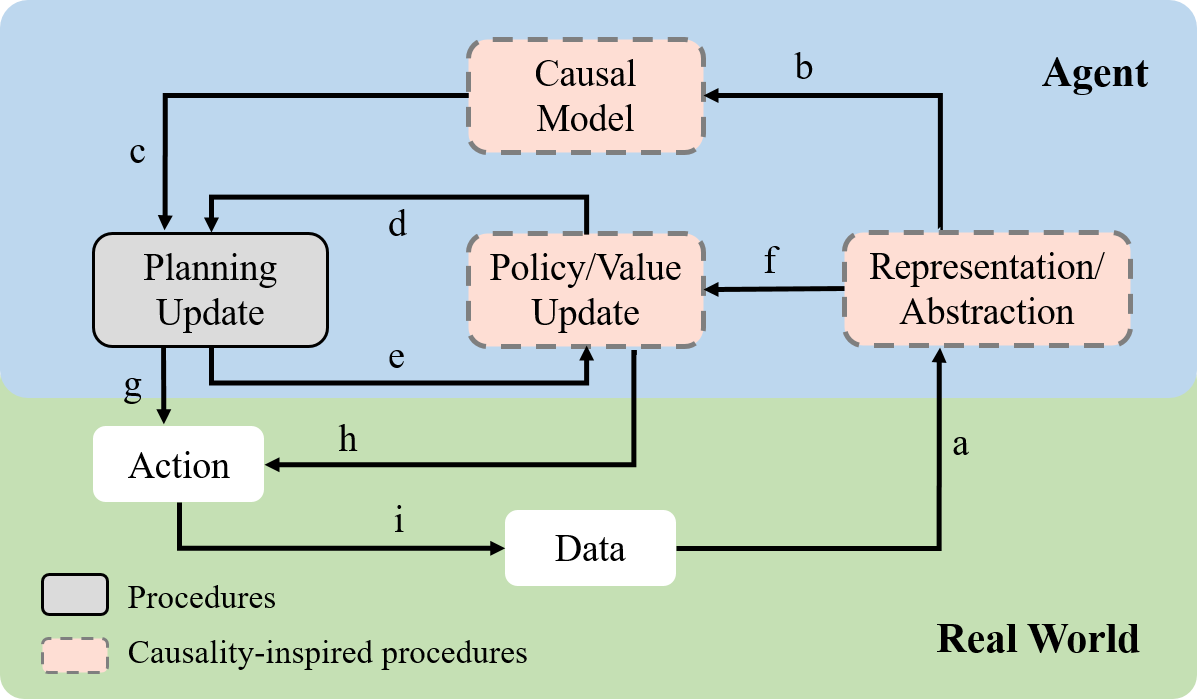}
    \caption{A sketch map of CRL framework, that illustrates how causality information inspires current RL algorithms. This framework contains possible algorithmic connections between planning and causality-inspired learning procedures. Explanations of each arrow are: a) input training data for the causal representation or abstraction learning; b) input representations, abstractions or training data from real world for the causal model; c) plan over a learned or given causal model, d) use information from a policy or value network to improve the planning procedure, e) use the result from planning as training targets for a policy or value, g) output an action in the real world from the planning, h) output an action in the real world from the policy/value function, f) input causal representations, abstractions or training data from real world for the policy or value update.}
    \label{fig:framework}
\end{figure}

The sketch map of CRL framework is shown in Fig.~\ref{fig:framework}, which gives an overview of possible algorithmic connections between planning and causality-inspired learning procedures. Causality-inspired learning could take place at three locations: in learning causal representations or abstractions (arrow a), learning a dynamics causal model (arrow b), and in learning a policy or value function (arrows e and f). Most CRL algorithms implement only a subset of the possible connections with causality, enjoying potential benefits in data efficiency, interpretability, robustness or generalization of the model or policy.

\begin{table*}[t]	
\label{tab:summary}
	\caption{Overview characteristics of existing CRL methods.}
	\begin{tabular}{llll} 
	    \toprule
		\multicolumn{1}{c}{Problems}  &\multicolumn{1}{c}{\bf CRL with Unknown Causal Information} &\multicolumn{1}{c}{\bf CRL with Known Causal Information} &\multicolumn{1}{c}{\bf Classical RL}\\ 
		\midrule
		Environment Modeling & compact causal graphs, confounding effects & compact causal graphs, confounding effects  & fully connected graphs \\
		Off-Policy Learning and Evaluation & action influences detection & confounding effects & no confounding effects\\
		Data Augmentation  & structural causal model learning & intervention and counterfactual reasoning & model-based RL \\
		Generalization  & causal discovery and causal invariance & invariance with causal information & invariance
		\\
		Theory Analysis  & causal identifiability and convergence\footnote{The convergence theory considers the sample complexity and regret bounds of the learned policy.} & convergence with causal information & convergence
		\\
		\bottomrule
	\end{tabular}
\end{table*}
According to different types of problems related to causality, we give an overview characteristics for CRL and traditional RL algorithms, with brief descriptions illustrated in Tab.\ref{tab:summary}. 
In particular, as for the environment modeling, CRL usually employs the structural equation model to represent relationships between variables of interest as a compact causal graph, and it may consider the latent confounders; while classical RL usually treats the environment model as a fully directed graph, e.g., all states at time $t$ would impact all states at time $(t+1)$.  
As for the off-policy learning and evaluation, CRL with unknown causal information may evaluate the influences of different actions, while that with known causal information usually investigates the confounding effects on the policy, by sensitivity analysis. Traditional RL would not model the confounding effects.
As for the data augmentation problem, classical RL sometimes is based on the model-based RL, while CRL on the structural causal model. After learning such a model, CRL could perform counterfactual reasoning to achieve data augmentation.
As for the generalization, classical RL attempts to explore invariance while CRL tries to leverage causal information to produce causal invariance, e.g., structural invariance, model invariance, etc.
As for the theoretical analysis, classical RL often concerns convergence, including the sample complexity and regret bounds of the learned policy, or the model errors; CRL also concerns convergence but with causal information, and it focuses on causal structure identifiability analysis.

\section{Evaluation Metrics}
In this section, we list some commonly used evaluation metrics in causal reinforcement learning, which essentially originate from fields of RL and causality. 

\subsection{Policy}
To evaluate the algorithm's performances, one often assesses the estimated policy via the average cumulative reward, average cumulative regret, fractional success rate, probability of choosing the optimal action, or their variants.
Cumulative rewards are averaged over episodes (or over runs) with the defined reward function in the training or testing phases~\cite{zhang2016markov,lu2018deconfounding,dasgupta2019causal,feliciano2021causal,liang2021inferring,feng2022factored,sun2021model}.
Cumulative regrets, i.e., the reward difference between the optimal policy and the agent’s policy, are computed and averaged over episodes (or over runs)~\cite{bareinboim2015bandits,lu2021efficient,li2021unifying,lu2020regret,tennenholtz2021bandits,subramanian2022causal}.
The Fractional Success Rate (FSR) is usually deployed in evaluation in robotic manipulation or related tasks, e.g., picking up an object to a target location.
FSR measures the overlapping success ratio between the object and the target~\cite{ahmed2020causalworld,zhu2021causaldyna}. 
In causal bandit problems, the probability of choosing the optimal arm can be used as the evaluation measurement~\cite{bareinboim2015bandits}. It measures the percentage of the optimal action over all time steps in each episode~\cite{lu2018deconfounding}.

\subsection{Model}
In causal reinforcement learning with the environment model, one enables to evaluate the quality of the learned dynamics. Such a quality is measured by the transition loss of the dynamics, between the ground truth states and predicted ones~\cite{wang2022causal,zhu2022invariant,zhang2020invariant}, e.g., 
\begin{equation*}
    Errors_{M} = \sum_{episodes} \sum_t (s_t - \hat{s}_t)^2,
\end{equation*}
where $s_t$ stand for ground truth states while $\hat{s}_t$ are the predicted ones from the dynamic model. Normally, a low transition loss helps learn a good policy.

\subsection{Causal Structure}
When identifying causal structures in causal reinforcement learning, one may evaluate the accuracy of the structure learning. As such, existing works often compute the edge accuracy (formally called precision) with the ground truth graph~\cite{zhu2022offline,wang2022causal}. In causal discovery domain, recall, and F1 values are employed for the evaluation as well. They are defined as below, 
\begin{gather*}
        Precision  = \frac{TP}{TP+FP}, \\
        Recall  = \frac{TP}{TP+FN}, \\
        F1 = \frac{2\times Precision\times Recall}{Precision+Recall},
\end{gather*}
where $TP, FP, FN$ are true positives, false positives, and false negatives, respectively. $TP$ correctly indicates the presence of an edge in the graph; $FP$ wrongly indicates that an edge is present while $FN$  wrongly indicates that an edge is absent in the graph.

\begin{figure*}
    \centering
    \includegraphics[scale=0.45]{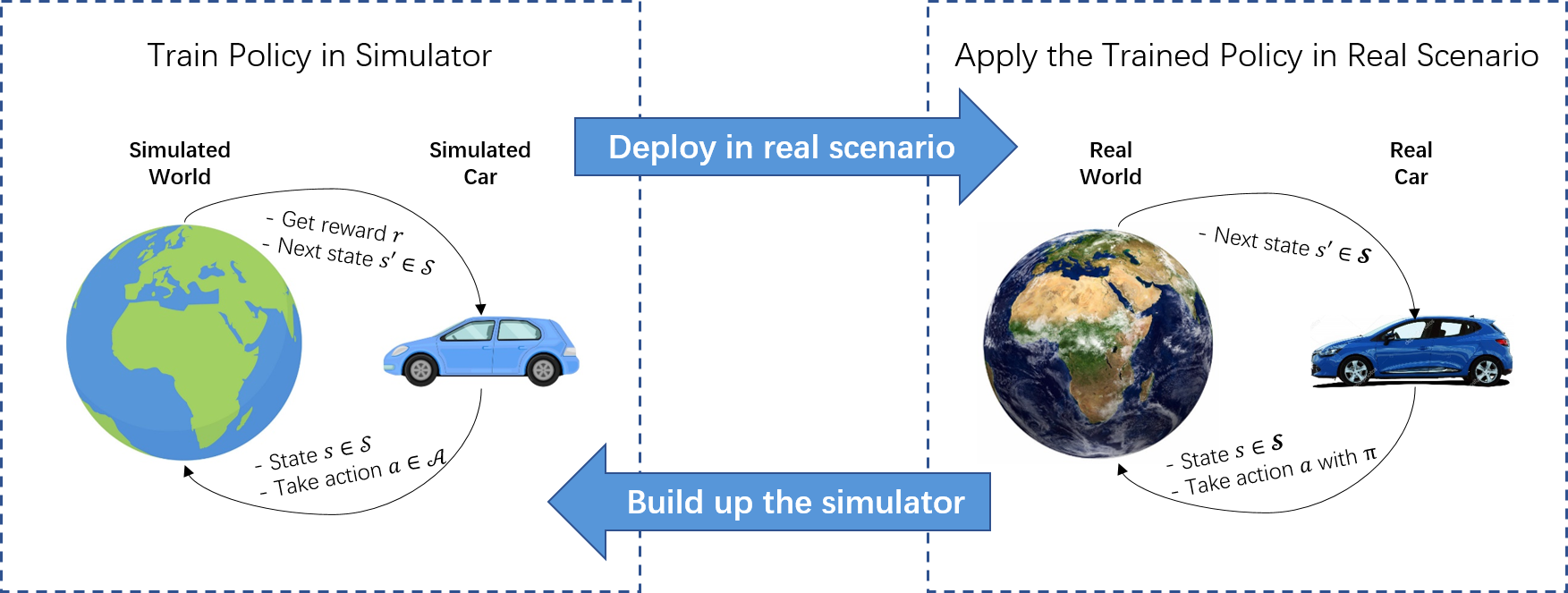}
    \caption{The overall framework of real world applications from CRL methods. On the one hand, a simulator is built up or corrected from the real scenario; while on the other hand, one may deploy the trained policy from simulator into real scenario.}
    \label{fig:simu2real}
\end{figure*}
\section{Applications}
Real world applications avoid errors from risky exploration and exploitation, while on the contrary CRL methods attain the trial-and-error mechanism.
To break up the contradicts between CRL and a specific real-world application, a credible strategy is to build up an authentic simulator based on some real data and domain knowledge about the model dynamics, then design the objectives for the agent and train the policy network in the simulator, and eventually deploy the trained policy in the real world with further improvement. The overall framework of real world applications from CRL methods is depicted in Fig.\ref{fig:simu2real}.  
Despite its fantastic and potential future, CRL still lies in infancy. Most current CRL methods basically study the performances in the simulator, with few of them in real-world scenarios. 
In this section, we review applications of the CRL methods in various fields, including but not limited by robotics, games, medical treatment, and autonomous driving, recommendation, etc., most of which are evaluated in simulators.

\subsection{Robotics}
Robotics has been applied flourishing in reinforcement learning~\cite{kober2013reinforcement,deisenroth2013survey,li2017deep}. And researchers has attempted to apply CRL as well in robotics to verify the performances of methods, since CRL can be incorporated in a straightforward way for robots to conduct some tasks via acting intellectually in complex open-world environments. 

Some prior works have explored causal IL robotic framework for learning and generalizing robotic skills~\cite{katz2016imitation}. These methods infer a hierarchical representation of a demonstrator's intentions, which is used to the planning procedure~\cite{katz2016imitation,katz2017novel,katz2017cognitive}.~\cite{ijcai2021-684} takes the best of the learned causal structure to foster policy learning for a given task of service robotics.~\cite{zhang2021learning} evaluates their causal dynamics and policy learning in the \textit{robosuite} simulation framework.~\cite{lee2021causal} focuses on robot manipulation tasks of block stacking and crate opening for policy transfer verification.~\cite{feng2022factored} proposes a factored adaptation algorithm for non-stationary RL, encoding the temporal changes of dynamics and rewards, and considers two robotic Sawyer manipulation tasks (Reaching and Peg) and a minitaur robot that learns to move at a target speed.~\cite{liang2020learning,liang2021inferring} discover time-delayed causal relations between events from a robot at regular or arbitrary times, while learning the optimal policy in the model-based RL framework. Methods are evaluated in two robotic experiments in the \textit{Gazebo} simulator, and a real robot for the painting task.~\cite{sonar2021invariant} proposes an invariant policy optimization algorithm and utilizes the colored-key problem for evaluation where the robot has to learn to navigate to the key, use it to open the door, and then navigate to the goal. They further consider a challenging task where the robot learns to open a door with different door's phisical parameters.
Other relative works with robotic manipulation applications include~\cite{pitis2020counterfactual,mozifian2020intervention} 

\begin{figure}
    \centering
    \includegraphics[scale=0.4]{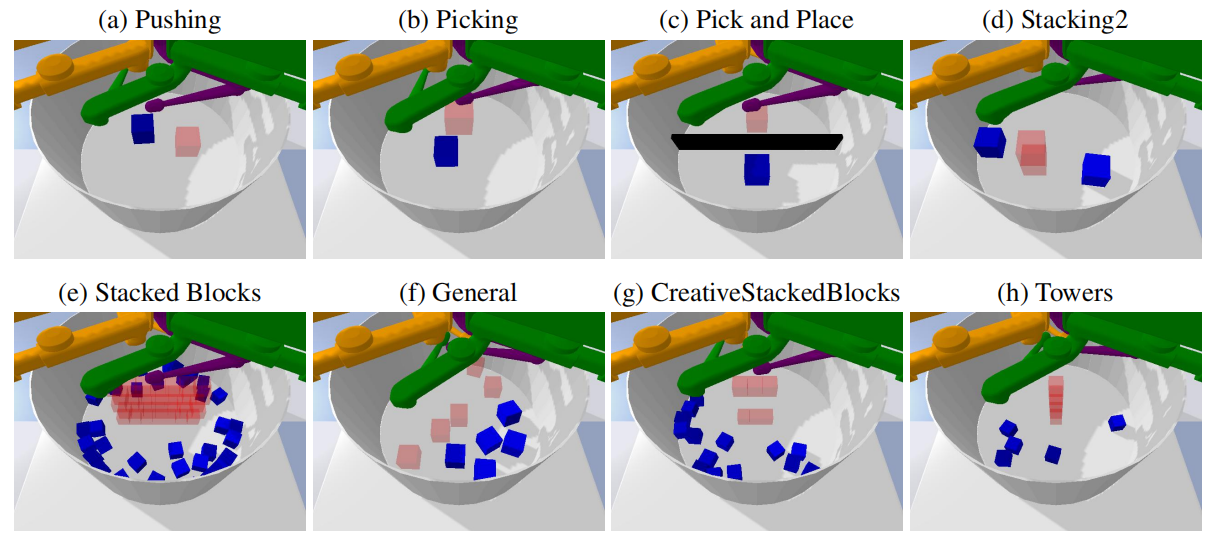}
    \caption{Some example tasks provided in the benchmark of \textit{CausalWorld} where the opaque red blocks stand for the goal shape while the blue ones are available blocks to be processed~\cite{ahmed2020causalworld}.}
    \label{fig:causalworld_example_tasks}
\end{figure}

Other prior works have explored robotic platforms for constructing a family of robotic manipulation environments, i.e., \textit{CausalWorld}~\cite{ahmed2020causalworld}. 
It is an open-source platform which allows for causal structure learning and transfer learning on new robotic environments. The robot in \textit{CausalWorld} is a 3-finger gripper, and each finger has three joints. The robot learns to move objects to specified target locations. Fig.~\ref{fig:causalworld_example_tasks} depicts some example tasks provided in the benchmark where the opaque red blocks stand for the goal shape while the blue ones are available blocks to be processed.
~\cite{sontakke2021causal,zhu2021causaldyna} use \textit{CausalWorld} to verify their methods' transfer learning and zero-shot generalizability in down-stream tasks.

For multi-joint dynamics control, one of the most widely applicable robotic platforms in RL belongs to MuJoCo. MuJoCo stands for Multi-Joint dynamics with Contact, which is a general-purpose physics engine consisting of 10 simulated environments of continuous controlling~\cite{todorov2012mujoco}. Researchers validate their approaches with improved performances of policy learning on MuJoCo-based Humanoid~\cite{tomar2021modelinvariant,chen2021instrumental}, Hopper~\cite{de2019causal}, Inverted Pendulum~\cite{zhu2022offline}, Half Cheetah~\cite{chen2021instrumental,swamy2022causal,zhang2020invariant}, Ant~\cite{swamy2022causal}, Walker~\cite{chen2021instrumental}, etc.

\subsection{Games}
Games enable to provide excellent testbeds for RL including CRL agents, since games, especially video games can vary enormously but still present challenging and interesting objectives for humans.

As for Atari games,~\cite{kansky2017schema} introduces Schema Networks to explore multiple causes of events for reasoning, resulting in high efficiency and zero-shot transfer on variations of game Breakout. Variations of such a game include standard version, versions with a middle wall, offset paddle, etc.~\cite{huang2022adarl} evaluates their proposed AdaRL algorithm that adapts to changes across domains, in the modified Atari Pong environments. They set changes in the orientations, colors, or reward functions in Pong to test robustness to change factors.~\cite{de2019causal} uses Atari games to state causal misidentification phenomenon indeed adversely influence imitation learning performance.
Works that utilize various Atari games to evaluate the performances, sample efficiency or generalization ability include~\cite{zhang2019learning,zhu2022invariant,pitis2020counterfactual}, etc.
As for Sequential Social Dilemmas (SSDs) which are spatially and temporally extended multi-agent games,~\cite{jaques2019social} chooses a public goods game \textit{Cleanup}, and a tragedy-of-the-commons game \textit{Harvest} to empirically demonstrate the enhancement of the proposed social influence reward. 
As for the StarCraft games,~\cite{foerster2018counterfactual} applies decentralized StarCraft micromanagement to evaluate their COMA method's efficacy, while~\cite{madumal2020explainable} adopts Starcraft II to evaluate the faithfulness of their proposed action influence models in explainability of behaviors.
Peng et al.,~\cite{peng2022causalitydriven} employ 2d-Minecraft~\cite{sohn2018hierarchical} and simplified
sandbox survival games Eden~\cite{chen2021eden} to verify the high quality of subgoal hierarchy in HRL driven by causality. 2d-Minecraft and Eden are two types of games with complex environments and especially with sparse rewards.
In addition, VizDoom, grid-world, and Box2D games have also been widely investigated in CRL for performance verification~\cite{zhang2019learning,volodin2020resolving,bennett2021off,kallus2020confounding,zhu2022invariant,huang2021action,swamy2022causal}, etc. 

\subsection{Healthcare}
Applying reinforcement learning related techniques in healthcare has been appealing but it also poses some significant challenges since it is concerned about one's safety and health~\cite{li2017deep,levine2020offline}. Performing online exploration on patients is largely impeded while some medical treatments might be costly, dangerous, or even unethical. Further, often only a few records of each patient for the healthcare data are available, and patients may give different responses to the same treatments. 

Regarding to heterogeneity across individuals,~\cite{lu2020sample} proposes to model the personalized treatment and general treatment, and investigates the method's performances on the real-world Medical Information Mart for Intensive Care-III (MIMIC-III) database with Sepsis-3. 
MIMIC-III is an influential dataset in healthcare treatment, which contains 6631 medical records from Intensive Care Units (ICUs)~\cite{johnson2016mimic}. This dataset has also been used by~\cite{bica2020learning,bica2021invariant}.
Outside of MIMIC-III dataset,~\cite{chen2021estimating} considers a
dataset with a total of 183, 487 mothers who delivered exactly two births during 1995 and 2009 in
the Commonwealth of Pennsylvania. Their aim is to estimate the DTR from retrospective
observational data using a time-varying instrumental variable.
~\cite{kallus2021minimax} focuses on a case study on the parallel Women’s
Health Initiative (WHI) clinical data to illustrate improvement of their confounded method in policy learning.~\cite{kallus2018confounding} studies the International Stroke Trial (IST), demonstrating their safety guarantees towards confounded selection.~\cite{zhang2019near,zhang2020designing} test the model of multi-stage treatment regimes for lung cancer or dyspnoea. 
Regarding to the safety and sample scarcity issues, other prior works construct a synthetic medical environment for performances investigation~\cite{gonzalez2018playing,tennenholtz2020off,killian2022counterfactually}.
%
~\cite{oberst2019counterfactual} applies a simple Sepsis simulator to construct a medical environment for illustration, while~\cite{namkoong2020off} uses simulators of Sepsis and autism management to certify the robustness of unobserved confounding.~\cite{shi2022off} performs a mobile health study where the synthetic dataset is generated by a generative model and from a mobile health application that monitors the blood glucose level of type 1 diabetic patients.~\cite{tennenholtz2022covariate} constructs an assistive-healthcare environment for the simulated robots to offer physical assistance to disabled persons.
%

\subsection{Autonomous Driving}
Autonomous driving is a particular suitable application in CRL, since safety and efficiency are two major barriers, and RL along with causality knowledge can be regarded as a potentially promising tool for enabling safe, effective learning in autonomous driving. As the widespread availability of high-quality demonstration data, imitation learning is a popular approach in autonomous driving~\cite{levine2020offline}.

Through the real-world driving setting, Codevilla et al.~\cite{codevilla2019exploring} and de Haan et al.~\cite{de2019causal} demonstrate the problems of causal misidentification and causal confusion in IL that accessing history or spurious correlations yields better validation performance, but worse actual driving performance. They overcome the problems by learning the correct causal model in the environment to find the correct policy. 
To verify the causal transfer ability for IL under the sensor shift, Etesami and Geiger~\cite{etesami2020causal} employ a real-world dataset that contains recordings of human-driven cars by drones that fly over several highway sections in Germany. Zhang et al.~\cite{zhang2020causal} test their causal imitation method using HighD dataset, which contains drone recordings of human-driven cars as well. 

Since most CRL has not yet been applied in actual real-world autonomous driving vehicles, simulated self-driving environments have been gaining increasing popularity, e.g., CARLA~\cite{codevilla2019exploring,zhang2021learning,samsami2021causal}, Car Racing~\cite{huang2021action}, Toy Car Driving~\cite{zhu2022offline}, etc. 
CARLA environment consists of different towns with different lengths of accessible roads and number of cars in urban areas. Apart from the original benchmark, Codevilla et al.~\cite{codevilla2019exploring} devise a larger scale CARLA benchmark, called NoCrash, to further evaluate their method's efficacy in complex conditions with dynamic objects. 
Zhang et al.~\cite{zhang2021learning} construct a highway driving scenario for evaluation.
Car Racing is a continuous task for a car with three continuous actions, and it has been used to validate the efficiency and efficacy of Action-Sufficient state Representations (ASRs) in improving policy learning~\cite{huang2021action}.
Zhu et al.~\cite{zhu2022offline} take the best of Toy Car Driving environment, which is a typical RL environment for a car to carry out tasks of avoiding obstacles or navigating, to evaluate the performances of causal structure learning. State variables include the car's direction, the velocity, and the position, etc.

\subsection{Recommendation}
Recommendation systems deal with the problem of recommending multiple items in sequence to the user on the basis of the user's delayed feedback~\cite{ye2019applying}. For instance, to pursue revenue maximization, online car hailing systems often recommend sequentially and adaptively different levels of riding coupons to passengers, according to their past car hailing behaviors and feedback.
Due to the difficulty for the simulator to offer fully observed environment information of the real world and the existence of unobserved confounding variables, CRL solutions could be appealing. 

In terms of the simulator of recommendation systems, Shi et al.~\cite{shi2019virtual} establish Virtual-Taobao for better commodity search in Taobao, which is one of the largest online retail platforms. Their simulator allows to train a recommendation policy with no physical sampling costs. In terms of the CRL approaches in recommendation systems, Shang et al.~\cite{shang2019environment} propose a deconfounded multi-agent environment
reconstruction (DEMER) method to achieve environment reconstruction with latent confounders. They apply an artificial application of driver program recommendation and a real application of Didi to verify the efficacy of DEMER.
Lu et al.~\cite{lu2020regret} use the email campaign data from Adobe, where the reward is obtained after the commercial links inside the email are clicked by the recipient. They demonstrate that their proposed bandit methods with causal information yield lower average regrets than those pure UCB or TS methods. Li et al.~\cite{li2021unifying} adopt Weixin A/B testing data with associated logs, as well as Yahoo’s news recommendation data, to show the superiority of using logged data and online feedbacks when making decisions. Tennenholtz et al.~\cite{tennenholtz2022covariate} conduct experiments to test the severity of covariate shifts in the confounded imitation setting on the RecSim Interest Evolution environment, which is a slate-based recommender system for simulating sequential user interaction. Shi et al.~\cite{shi2022off} present an off-policy value estimation algorithm in the confounded MDP setting, and they justify their method with a real-world dataset from a ride hailing company. This dataset is about a recommendation program applied to customers in regions where there are more taxi drivers than the call orders, with which they aim at evaluating the coupon recommendation policies.

\subsection{Others}
RL has many other applications though, we here list existing works on CRL in the domain of classical control. Such domains as financial economy, natural language process, etc. are in need of exploration in CRL. 

In terms of classical control, Sun et al.~\cite{sun2021model} perform experiments on Pendulum, Mountain Car, and Cart Pole environments and simulated quadcopter recordings to verify the robustness of their framework. Zhang et al.~\cite{zhang2020invariant} give an illustration of the benchmark Cartpole and show they enable to learn Model-Irrelevance
State Abstractions (MISA) while training a policy. Chen et al.~\cite{chen2021instrumental} modify environments of Mountain Car, Cart Pole, and Catch to demonstrate how IV variables could address the confounding bias. De Haan et al.~\cite{de2019causal} create a variant of Mountain Car confounded testbed to indicate the necessity of disentanglement of the state representation, while Shi et al.~\cite{shi2022minimax} consider the Cart Pole environment for evaluation.

\section{Open Sources}

\subsection{Codes}
We summarize codes with prior causal information and unknown causal information in Tab.~\ref{tab:summary_known_codes} and Tab.~\ref{tab:summary_unknown_codes}, respectively.
\begin{table}[ht]	
	\caption{Part of CRL codes with prior causal information}
	\begin{center} 
	\begin{tabular}{  m{2cm} m{6cm} } 
	    \toprule
		\multicolumn{1}{c}{\bf Methods}  &\multicolumn{1}{c}{\bf Sources} \\ 
        \midrule
        Causal TS~\cite{bareinboim2015bandits} & \url{ https://github.com/ucla-csl/mabuc} \\
        \midrule
        IVOPE~\cite{chen2022instrumental} & \url{https://github.com/liyuan9988/IVOPEwithACME}\\
        \midrule
        DFPV~\cite{xu2021deep} & \url{https://github.com/liyuan9988/DeepFeatureProxyVariable/} \\
        \midrule
        COPE~\cite{shi2022off} & \url{https://github.com/Mamba413/cope}\\
        \midrule
        off\_policy\_con founding \cite{namkoong2020off}  & \url{https://github.com/StanfordAI4HI/off_policy_confounding} \\
        \midrule
        RIA~\cite{guo2022relational} & \url{https://github.com/CR-Gjx/RIA} \\
        \midrule
        \item Gumble-Max SCM~\cite{oberst2019counterfactual} & \url{https://github.com/clinicalml/gumbel-max-scm} \\
        \midrule
        Confounded-POMDP~\cite{shi2022minimax}  & \url{https://github.com/jiaweihhuang/Confounded-POMDP-Exp} \\
        \midrule
        Causal Bandit~\cite{lattimore2016causal} & \url{https://github.com/finnhacks42/causal_bandits} \\
        \midrule
        POMIS~\cite{lee2018where} &  \url{https://causalai.net/} \\
        \midrule
        Unc\_CCB~\cite{subramanian2022causal} & \url{https://openreview.net/forum?id=F5Em8ASCosV} \\
        \midrule
        OptZ~\cite{bennett2019policy} & \url{https://github.com/CausalML/LatentConfounderBalancing} \\
        \midrule
        CRLogit~\cite{kallus2018confounding} & \url{https://github.com/CausalML/confounding-robust-policy-improvement} \\
        \midrule
        CTS~\cite{tennenholtz2022covariate} & \url{https://openreview.net/forum?id=w01vBAcewNX} \\
        \midrule
        ResidualIL~\cite{swamy2022causal} &  \url{https://github.com/gkswamy98/causal_il}\\
		\bottomrule
  \label{tab:summary_known_codes}
	\end{tabular}
	\end{center}
\end{table}
\begin{table}[ht]	
	\caption{Part of CRL codes with unknown causal information}
	\begin{center} 
	\begin{tabular}{  m{2cm} m{6cm} } 
	    \toprule
		\multicolumn{1}{c}{\bf Methods}  &\multicolumn{1}{c}{\bf Sources} \\ 
         \midrule
        CIM~\cite{samsami2021causal} & \url{https://github.com/vita-epfl/CIM} \\
        \midrule
        ICIN~\cite{nair2019causal} & \url{https://github.com/StanfordVL/causal_induction}\\
        \midrule
        causal InfoGAN~\cite{kurutach2018learning} & \url{https://github.com/thanard/causal-infogan} \\
        \midrule
        NIE~\cite{Herlau2022reinforcement} & \url{https://gitlab.compute.dtu.dk/tuhe/causal_nie} \\
	\midrule
        DBC~\cite{zhang2021learning}  & \url{https://github.com/facebookresearch/deep_bisim4control} \\
        \midrule
        IBIT~\cite{mozifian2020intervention} & \url{https://github.com/melfm/ibit} \\
        \midrule
        CoDA~\cite{pitis2020counterfactual} & \url{https://github.com/spitis/mrl} \\
        \midrule
        CAI~\cite{seitzer2021causal} & \url{https://github.com/martius-lab/cid-in-rl} \\
        \midrule
        CDL~\cite{wang2022causal} & \url{https://github.com/wangzizhao/CausalDynamicsLearning} \\
        \midrule
        Deconf. AC~\cite{lu2018deconfounding} &            \url{https://github.com/CausalRL} \\
        \midrule
        AdaRL~\cite{huang2022adarl}  & \url{https://github.com/Adaptive-RL/AdaRL-code} \\
        \midrule
        IPO~\cite{sonar2021invariant} & \url{https://github.com/irom-lab/Invariant-Policy-Optimization} \\
        \midrule
        linear CB~\cite{tennenholtz2021bandits} & \url{https://github.com/guytenn/Bandits_with_Partially_Observable_Offline_Data} \\
        \midrule
        causal misidentification~\cite{de2019causal}  & \url{https://github.com/pimdh/causal-confusion}.\\
        \midrule
        CILRS~\cite{codevilla2019exploring} & \url{https://github.com/felipecode/coiltraine/blob/master/docs/exploring_limitations.md}\\
        \midrule
        ICIL~\cite{bica2021invariant} & \url{https://github.com/ioanabica/Invariant-Causal-Imitation-Learning} \\
        \midrule
        cause-effect IL~\cite{katz2017novel} & \url{https://github.com/garrettkatz/copct} \\
        \midrule
        fighting-copycat~\cite{wen2020fighting} & \url{https://github.com/AlvinWen428/fighting-copycat-agents} \\
        \midrule
        keyframe~\cite{wen2021keyframe} & \url{https://tinyurl.com/imitation-keyframes} \\
        \midrule
        PrimeNet~\cite{wen2022fighting} & \url{https://github.com/AlvinWen428/fighting-fire-with-fire}\\
        \midrule
        causal-rl~\cite{gasse2021causal} & \url{https://github.com/causal-rl-anonymous/causal-rl} \\
		\bottomrule
  \label{tab:summary_unknown_codes}
	\end{tabular}
	\end{center}
\end{table}

\subsection{Tutorials, Reports or Surveys}
One of the most informative and comprehensive tutorials belongs to the one presented by Bareinboim~\cite{Elias2020icml} at ICML 2020:~\url{https://crl.causalai.net/}. He introduced the foundations of causal inference and reinforcement learning as well as discussed six fundamental and prominent CRL tasks. Such tasks are i) generalized policy learning; ii) when and where to intervene; iii) counterfactual decision making; iv) generalizabililty and robustness of causal claims; v) learning casual models; and vi) casual imitation learning. 
Other technical reports, blogs or surveys are shown as below:
\begin{itemize}
    \item Lu~\cite{lu2018introduction} was motivated by the current development in healthcare and medicine, and introduced causal reinforcement learning. He summarized the advantages of CRL as \textit{data efficiency} and \textit{minimal change}.
    \item A company called {causalLens} briefly summarized current research trend of CRL in cases where there are latent confounders or in deep model-based CRL:~\url{https://www.causalens.com/technical-report/causal-reinforcement-learning/}.
    \item Ness shared his bet on CRL as the next killer application of marketing science within the next ten years:~\url{https://towardsdatascience.com/my-bet-on-causal-reinforcement-learning-d94fc9b37466}.
    \item Grimbly et al.,~\cite{grimbly2021causal} surveyed on causal multi-agent reinforcement learning. Grimbly also discussed the six tasks from~\cite{Elias2020icml} with surveying relevant CRL algorithms:~\url{https://stjohngrimbly.com/causal-reinforcement-learning/}.    
    \item Bannon et al.,~\cite{bannon2020causality} surveyed on causal effects' estimation and off-policy evaluation in batch reinforcement learning.
    \item Kaddour et al.,~\cite{kaddour2022causal} surveyed on causal machine learning that included a chapter of CRL, summarizing causal bandits algorithms, etc. 
\end{itemize}

\subsection{Competitions}
Apart from the fruitful competitions with reinforcement learning that CRL algorithms could be exploited, there exists another competition related to both causality and reinforcement learning.
That is the Learning By Doing: Controlling a Dynamical System using Control Theory, Reinforcement Learning, or Causality. It is a competition of NeurIPS 2021, which aims at encouraging cross-disciplinary discussions among researchers from fields of control theory, RL and causality. 
Two tracks were designed.
In Track CHEM, the goal is to find optimal policies for the concentrations of chemical compounds to ensure a specific target concentration on one of the products. 
In Track ROBO, the goal is to learn optimal policies for a robot arm, such that the tip of the robot follows a target process via sequentially interaction. 
For detailed background information or solutions, please refer to:~\url{https://learningbydoingcompetition.github.io/}~\cite{weichwald2022learning}.

\section{Future Directions}
We categorize current CRL approaches into two sides, according to whether their causal information is given or not.
While it is not that easy to attain and take the best of causal information from pure observed data, in this section, we discuss some future directions in CRL from the following 4 aspects: i) knowledge-based: how to extract knowledge when causal information is unknown, ii) memory-based: how to store and remember causal information, iii) cognition-based: how to exploit prior causal information from related tasks or experts, and iv) application-based: how to apply CRL algorithms in real worlds.

\subsection{Knowledge-based causal reinforcement learning}
Knowledge-based CRL aims at incorporating knowledge into CRL, and often involves the procedure of extracting knowledge from data. When encountering new environments or tasks, updating knowledge is more than needed. Knowledge may have various forms (e.g., causal structure, causal ordering, causal effects, causal features, etc.) and may be incorporated in the model, value function, reward function, etc. 

Taking the environment model for instance, how to ensure that the estimated environment model with knowledge is consistent with the actual real-world one becomes a core challenge (distributional shift problem)~\footnote{Existing RL algorithms, to deal with the distributional shift problem, are mainly via policy constraints or uncertainty estimation~\cite{levine2020offline}.}.
To handle this challenge, one possible solution is to consider latent confounders in CRL. 
Since there is a great possibility for the existence of latent confounders in many applications, studying confounders enables current CRL methods to learn a more accurate model or policy, which mitigates the shortcomings from the differences between the learned policy and the optimal policy.  
Though there exist some works on confounders in CRL that we review in Section~\ref{sec:crl}, it is suggested that developing general methods with weaker assumptions to detect or deploy latent confounders and eliminate their effects, may be potentially relevant for effective, robust, and interpretable causal reinforcement learning.
%
Another possible solution to distributional shift is to take into account the non-stationarity of causal structures in the model, e.g., the causal relations between states, actions, and rewards. Such non-stationarity may be originated across multiple time steps, multiple domains, multiple modals of data, multiple tasks or even multiple agents, which implies that causal structures in different time steps, domains, modals, tasks, or agents may change. 
For instance, in robotic manipulation environments~\cite{ahmed2020causalworld}, the causal structure would change whenever a robotic arm touches or picks one of the objects; For the Pushing and Picking tasks, causal relations between actions and rewards are non-stationary; 
For the identical task in the same environment, different types of robotic arms (agents) may own different characteristics, rendering the same action by arms towards the same object possibly produce different dynamics, rewards or policy generation mechanisms; There also may exist cooperative, defensive or adversarial agents in the multi-agent setting.
However, the questions of how such non-stationarity of the knowledge affects the dynamics, rewards or policy functions, what real effects are and how to deal with them, remain a large region to be explored.

\subsection{Memory-based causal reinforcement learning}
Memory-based CRL expects to be possessed of powerful memory that preserves the sequences or the distributions of state, action as well as reward, so as to remember previously-visited regions (e.g., interesting states, actions, etc.) and
eventually achieve more effective and efficient policy learning in an intelligent design. Current RL algorithms, based on recurrent neural networks to integrate information over long time periods, usually deploy attention or differentiable memory techniques to store and retrieve successful experiences for rapid learning~\cite{arulkumaran2017brief,pritzel2017neural}.

To spawn memory-based CRL methods, one may gain advantages from the combination of both on-policy and off-policy algorithms, while avoiding getting stuck in local optima and improving convergence rates.
To keep memory and prevent forgetting those skills that ever trained, how to design an experience replay scheme to take the best of memory information for transfer learning, deserves exploration in CRL. Alternatively, the setting of lifelong CRL is also another interesting research direction to prevent forgetting based on memory replay.

\subsection{Cognition-based causal reinforcement learning}
Cognition-based CRL attempts to exploit previously acquired or newly updated cognition capabilities from related tasks or experts, to speed up learning in CRL with high performances. It follows the logic from the brain, and involves the processes of perception, understanding, judgment, evaluation, abstract thinking and reasoning, etc~\cite{enwiki}. 

Intervention and counterfactual analysis for CRL are recent emerging paradigms that utilize the existing cognitive model to improve sample efficiency and facilitate learning. 
Regarding intervention issue, with causal models taken into account in the sequential decision-making setting, the interesting avenues for research include questions of when, where and how to intervene variables of interests to lead to optimal policies, based on perception with less assumptions. These ways could be used as a heuristic for a possible refinement of the policy space based on topological constraints~\cite{Elias2020icml,lee2020characterizing,lee2019structural}, as well as for a renewal of the causal model by correcting the causal direction or causal knowledge. 
Regarding counterfactual analysis issue, since it involves thinking about possibilities or imagination in the cognitive process, it plays vital roles in investigating the outcomes from some dangerous, unethical or even impossible action spaces in safety critical applications. Causality offers tools for modeling such uncertain scenarios for analysis, thus enjoying sample efficiency and policy effectiveness.

A key to the success of cognition-based CRL belongs to the generalization capability of methods, including learning generalizable models as well as training generalizable policies towards other unseen testing environments, tasks or agents. 
Ahmed et al.,~\cite{ahmed2020causalworld} empirically find that in \textit{CausalWorld}, an agent’s generalization capabilities are related to the experience gathered at the training phase.
Based on the training experience, how to characterize the causal structural invariances and the variances that are shared or specific across domains, tasks or agents from the acquired knowledge, is still a potentially fruitful research direction. In addition to the structural invariances, invariant causal features or state abstractions may also be effective alternatives to aid in generalizability. Relevant techniques are meta-learning, domain generalization, etc. Besides, causal models open room for hierarchical RL to adaptively form the high-level policy instead of manually dividing the goals, which is an interesting area of ongoing research for generalization capability.

\subsection{Application-based causal reinforcement learning}
Application-based CRL focuses on applying current CRL algorithms in real-world applications such that the key requirements and challenges in the industrial community could be tackled. Possible applications include games, robotics (e.g., service robots, medical robots, robotic arms, aerial vehicle), military, healthcare (e.g., medical treatments), financial trading, product recommendation, urban transportation (e.g., autonomous driving), smart cities (e.g., electricity allocation), etc. Taking autonomous driving for instance, a car is required to reach the goal from the start safely and reliably, obeying the traffic rules with no accidents and enjoying real-time performances. 
To facilitate real-world applications of CRL, causal information from the simulator (or environment dynamics) or datasets, whether from prior knowledge or not, lacks investigation, which can be further explored in future work.

Building hand-crafted simulators has been widely adopted in RL, such as autonomous driving~\cite{Dosovitskiy17,zhou2021smarts}, robotic control~\cite{ahmed2020causalworld}, etc.. Simulators are beneficial to generate specific scenarios in real-world applications that are rare for robust policy training~\cite{luo2022survey}, e.g., attack-defense game or cooperative-adversarial scenarios in multi-agent settings, etc. However, since it is hard for simulators to obtain high fidelity compared with real-world scenarios, performances may be unreliable and sensitive to the simulator-reality gap when utilizing the trained policy network into real tasks. Thus, accompanied with the simulators, we have to look back to the principal overall functionality and demands from each of the applications to reduce as much simulator-reality gap as possible; accompanied with the algorithms, we could consider the techniques for generalization capability in cognitive-based CRL, to reduce the unfavorable effects from the gap and to improve the 
human-machine interaction capability.

While current CRL works place considerable value on design of simulators, new algorithms and theory, much practical progress in real-world applications might have been and will be driven by advances in datasets, similar to the fields of computer vision and natural language process. Collecting representative, large, and diverse datasets is often far more significant than utilizing the most advanced methods if factoring in real-world applications~\cite{levine2020offline}. Whereas most standard RL methods belong to interactive learning with the environment, collecting large and diverse datasets is often impractical, costly, or unethical, due to safety or privacy concerns. The machine learning and CRL communities shed light on the facts that i) Applications could well benefit from the real-world, large and diverse datasets; ii) Methods integrated with offline data enable to well handle a wide range of real-world problems. Hence, careful collections of available datasets, as well as the development of data-driven or data-fusion CRL algorithms, could potentially induce a series of breakthroughs towards more efficient, effective and practical CRL techniques in the near future. 

\section{Conclusions}
We have witnessed the amazing and flourishing progress of causal reinforcement learning in recent years.
In this survey, we take a comprehensive review on causal reinforcement learning, where existing CRL methods roughly fall into two categories. One category is based on the unknown causality information which has to learn before training policies, while the other one relies on the known causality information. For each category, we analyze the reviewed methods according to their model settings, i.e., settings of MDP, POMDP, Bandits, DTR or IL. 
We further offer evaluation metrics, open sources and some occurring applications in CRL. 
We eventually summarize future directions from the following four aspects: i) knowledge-based; ii) memory-based; iii) cognition-based; and iv) application-based CRL, which may empower CRL with a new era of advances.

\section*{Acknowledgments}
This work was supported in part by the National Key Research and Development Program of China under Grant 2021ZD0111501, in part by the National Science Fund for Excellent Young Scholars under Grant 62122022, in part by the Natural Science Foundation of China under Grant 61876043 and Grant 61976052, in part by the Guangdong Provincial Science and Technology Innovation Strategy Fund under Grant 2019B121203012, and in part by China Postdoctoral Science Foundation 2022M711812.

\bibliographystyle{IEEEtran}
\bibliography{mybibfile}

\newpage
\begin{IEEEbiography}[{\includegraphics[width=1in,height=1.25in,clip,keepaspectratio]{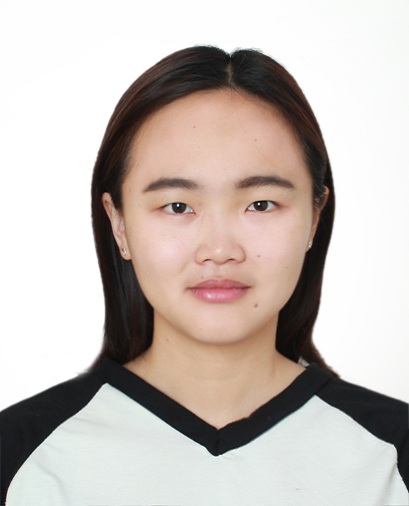}}]{Yan Zeng}
received her B.S. degree in the School of Applied Mathematics in 2016 and Ph.D. degree in the School of Computer at Guangdong University of Technology in 2021. She is currently a Postdoctoral Researcher in the Department of Computer Science and Technology, Tsinghua University. She was an intern in the Causal Inference Team, RIKEN Center for Advanced Intelligence Project, Tokyo, Japan from 2019 to 2020. Her current research interests include causal reinforcement learning, causal discovery, and latent variable modeling, etc.
\end{IEEEbiography}

\vspace{-11pt}
\begin{IEEEbiography}[{\includegraphics[width=1in, height=1.25in, clip, keepaspectratio]{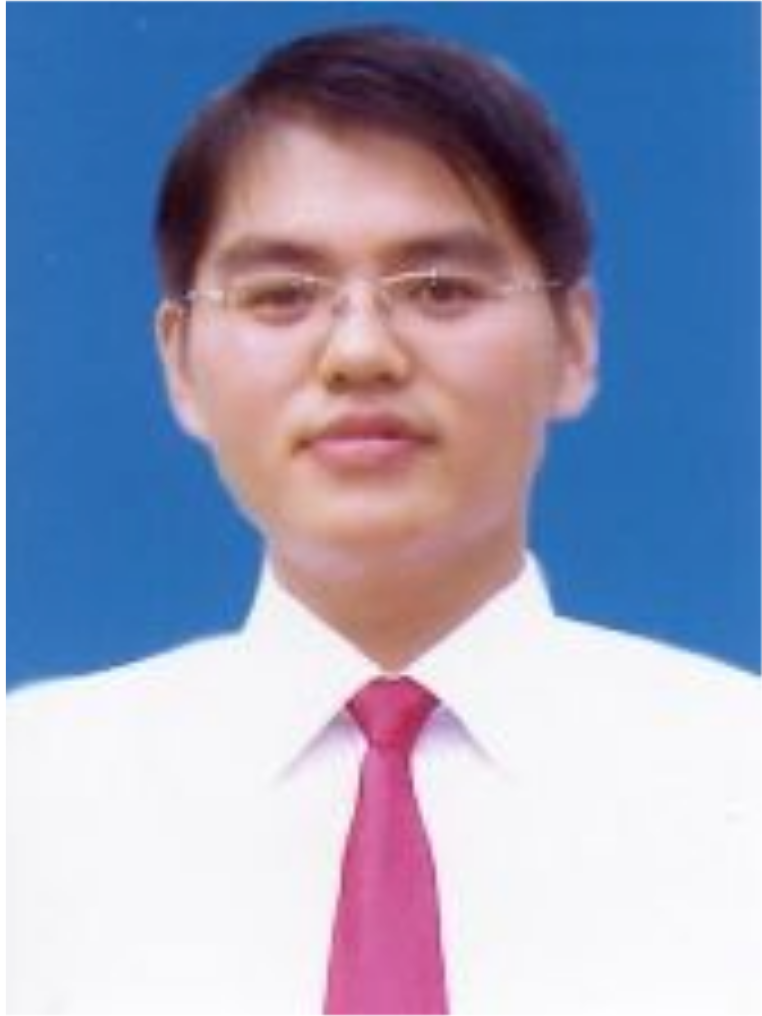}}]{Ruichu Cai}received his B.S. in Applied Mathematics and Ph.D. in Computer Science from South China University of Technology in 2005 and 2010, respectively. He is currently a Professor in the School of Computer, Guangdong University of Technology and in the Pazhou Lab., Guangzhou. He was a visiting student at National University of Singapore in 2007-2009, and research fellow at the Advanced Digital Sciences Center, Illinois at Singapore in 2013-2014. His research interests cover a variety of different topics including causality, machine learning and their applications.
\end{IEEEbiography}

\vspace{-11pt}
\begin{IEEEbiography}[{\includegraphics[width=1in, height=1.25in, clip, keepaspectratio]{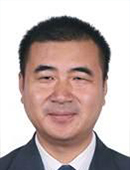}}]{Fuchun Sun}, IEEE Fellow, received the Ph.D. degree in computer science and technology from Tsinghua University, Beijing, China, in 1997. He is currently a Professor with the Department of Computer Science and Technology and President of
Academic Committee of the Department, Tsinghua University, deputy director of State Key Lab. of Intelligent Technology and Systems, Beijing, China. His research interests include artificial intelligence, intelligent control and robotics, information sensing and processing in artificial cognitive systems, etc. He was recognized as a Distinguished Young Scholar in 2006 by the Natural Science Foundation of China. He became a member of the Technical Committee on Intelligent Control of the IEEE Control System Society in 2006.
He serves as Editor-in-Chief of International Journal on Cognitive Computation and Systems, and an Associate Editor for a series of international journals including the IEEE TRANSACTIONS ON COGNITIVE AND DEVELOPMENTAL
SYSTEMS, the IEEE TRANSACTIONS ON FUZZY SYSTEMS, and the IEEE
TRANSACTIONS ON SYSTEMS, MAN, AND CYBERNETICS: SYSTEMS.
\end{IEEEbiography}

\vspace{-11pt}
\begin{IEEEbiography}
	[{\includegraphics[width=1in, height=1.25in, clip, keepaspectratio]{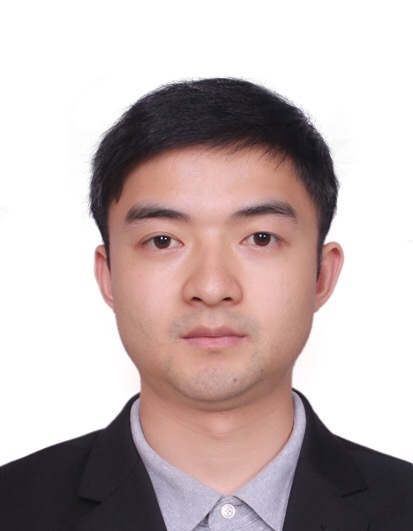}}]{Libo Huang}received his B.S. degree in Mathematics at Jiangxi Normal University in 2016 and his Ph.D. in the School of Information Engineering at Guangdong University of Technology in 2021. He was a joint Ph.D. student in the Department of Electronics and Computer Engineering at Brunel University London from 2019 to 2020. He is currently a Postdoctoral Researcher at the Institute of Computing Technology, Chinese Academy of Sciences. His research interests include deep continuous learning and causal reinforcement learning.
\end{IEEEbiography}

\vspace{-11pt}
\begin{IEEEbiography}
	[{\includegraphics[width=1in, height=1.25in, clip, keepaspectratio]{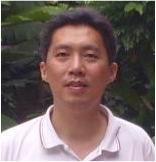}}]{Zhifeng Hao}received the B.S. degree in Mathematics from the Sun Yat-Sen University in 1990, and the Ph.D. degree in Mathematics from Nanjing University in 1995. He is currently a Professor in the College of Science, Shantou University. His research interests involve various aspects of Algebra, Machine Learning, Data Mining, Evolutionary Algorithms.
\end{IEEEbiography}


\end{document}